\documentclass[twoside]{article}

\usepackage[accepted]{aistats2026}

\usepackage[round]{natbib}

\usepackage{xcolor}
\newcommand{\red}[1]{\textcolor{red}{#1}}
\newcommand{\green}[1]{\textcolor{teal}{#1}}
\usepackage{amsmath}
\usepackage{amssymb}
\usepackage{enumitem}
\usepackage{longtable}

\usepackage{hyperref}
\usepackage{url}
\usepackage{theorem}

\theoremstyle{plain}
\newtheorem{theorem}{Theorem}[section]

\newtheorem{lemma}[theorem]{Lemma}

\theoremstyle{definition}

\newtheorem{assumption}[theorem]{Assumption}
\theoremstyle{remark}

\usepackage{graphicx}
\usepackage{epstopdf}
\usepackage{todonotes}
\newcommand{\rep}{\phi}

\newcommand{\binvar}{T}
\newcommand{\Pall}{P}
\newcommand{\Ps}{P^0}
\newcommand{\Pt}{P^1}
\newcommand{\E}{\mathbb{E}}

\newcommand{\EP}[1]{\mathbb{E}_{#1}}
\newcommand{\Es}{\EP{\Ps}}
\newcommand{\Et}{\EP{\Pt}}

\newcommand{\density}[3]{\frac{d{#3_{#1}}}{d{#2_{#1}}}}
\newcommand{\densityst}[1]{\density{#1}{\Ps}{\Pt}}

\newcommand{\misoverlapst}{\mathcal{O}}
\allowdisplaybreaks

\makeatletter
\def\@copyrightspace{%
\@float{copyrightbox}[b]
\begin{center}
\setlength{\unitlength}{1pc}
\begin{picture}(20,4.6)%
  \put(0,5.6){\line(1,0){4.818}}%
  \put(0,0){\parbox[b]{19.75pc}{\small \Notice@String}}%
\end{picture}
\end{center}
\end@float}
\makeatother

\begin{document}

\twocolumn[

\aistatstitle{Deconfounding Scores and Representation Learning for Causal Effect Estimation with Weak Overlap}

\runningauthor{Oscar Clivio, Alexander D'Amour, Alexander Franks, David Bruns-Smith, Chris Holmes, Avi Feller}

\aistatsauthor{Oscar Clivio\textsuperscript{*} \And Alexander D'Amour\textsuperscript{*} \And Alexander Franks\textsuperscript{*}}
\aistatsaddress{ University of Oxford \And Google DeepMind \And University of California, Santa Barbara}

\aistatsauthor{David Bruns-Smith \And Chris Holmes \And Avi Feller}
\aistatsaddress{Stanford University \And University of Oxford \\ Ellison Institute of Technology \And University of California, Berkeley}

]

\makeatletter
\renewcommand{\Notice@String}{\textsuperscript{*}Equal contribution.\\ \par\AISTATS@appearing}
\makeatother

\begin{abstract}
Overlap, also known as positivity, is a key condition for causal treatment effect estimation. Many popular estimators suffer from high variance and become brittle when features differ strongly across treatment groups. This is especially challenging in high dimensions: the curse of dimensionality can make overlap implausible. To address this, we propose a class of feature representations called \textit{deconfounding scores}, which preserve both identification and the target of estimation; the classical propensity and prognostic scores are two special cases. We characterize the problem of finding a representation with better overlap as minimizing an \textit{overlap divergence} under a deconfounding score constraint. We then derive closed-form expressions for a class of deconfounding scores under a broad family of generalized linear models with Gaussian features and show that prognostic scores are overlap-optimal within this class. We conduct extensive experiments to assess this behavior empirically.
\end{abstract}

\section{INTRODUCTION}

In observational causal inference, researchers typically emphasize the key role of the ignorability assumption, which assumes that observed features contain all confounders of the relationship between the treatment and the outcome. Equally important---yet less often the focus---is \textit{overlap} (or \textit{positivity}): the distributions of features in the treated and control groups must have common support. Both assumptions are critical for nonparametric identification, enabling flexible confounder adjustment for unbiased estimation of causal estimands.

Even when overlap holds, features can still differ substantially between treatment and control groups, degrading both theoretical guarantees and practical performance of importance weighting and doubly robust estimators \citep{rothe2017rcifateulo, hong2020iofpteulo}.
The problem is compounded in modern settings where adjusting for a large number of features is necessary for ignorability to be plausible \citep{damour2021oioswhdc}.
\emph{Representation learning} offers a natural strategy: learn a low-dimensional mapping of the features that preserves unbiased estimation while potentially improving overlap.
Two widely-used representations are \emph{balancing scores} (including \emph{propensity scores}) \citep{rosenbaum1983tcrotpsiosfce} and \emph{prognostic scores} \citep{hansen2008tpaotps}, which capture all feature information predictive of the treatment assignment and the outcome, respectively.

In this paper, we introduce \emph{deconfounding scores}: the complete class of representations that preserve unbiased estimation of the target estimand under ignorability. Balancing and prognostic scores are special cases. Our main contributions are:
\begin{itemize}
\item We characterize the class of deconfounding scores and specify the conditions necessary to compute them; this class naturally falls on a continuum between prognostic and balancing scores.

\item We introduce an \textit{overlap divergence} that quantifies the lack of overlap with respect to a representation and show that it controls the semiparametric efficiency bound.

\item We establish that (i) representations always improve overlap---as measured by overlap divergence---compared to original covariates, and (ii) those that are less predictive of the treatment assignment exhibit better overlap.

\item In the canonical setting with Gaussian features and generalized linear models for the outcome and treatment assignment, we analytically characterize a family of deconfounding scores as lying on a hyperbola whose endpoints correspond to balancing and prognostic scores. For this family, \textbf{prognostic scores minimize the overlap divergence while balancing scores maximize it}.

\item In simulations, a correctly-specified prognostic score improves performance over raw covariates and most other deconfounding scores. On semi-synthetic datasets, there is always at least one deconfounding score improving over covariates, with the prognostic score yielding the most improvements over covariates.
\end{itemize}

\subsection{Related Work}

\paragraph{Inference with Poor Overlap.} When the overlap assumption is not satisfied, the conditional average treatment effect (CATE) is no longer nonparametrically identified on points or regions without overlap \citep{hernan2010ci}. This can be mitigated with the help of assumptions on the outcome or propensity models, by extrapolating the CATE from regions with overlap to those without overlap \citep{petersen2012dartvitpa, nethery2019epaceitponoteongcseocm}, or through \emph{partial identification}, that is, bounds on the CATE on regions without overlap \citep{manski1990nbote, lee2021btebpliao, khan2024opebopits}. An alternative is the use of balancing weights to estimate population-wide treatment effects \citep{kallus2020gommfci, brunssmith2022oaadtfbw}. Even when overlap holds, stricter versions are required for root-$n$ confidence intervals  \citep{khan2010iiscaiwe, rothe2017rcifateulo, hong2020iofpteulo} but are again often unrealistic in high dimensions \citep{damour2021oioswhdc}. To tackle this, previous approaches have typically focused on changing the treatment effect estimand to a population with better overlap  \citep{crump2009dwloieoate, matsouaka2020affciitpoeipwtroow}, trimming extreme estimated propensity scores \citep{sturmer2010teitpoucdwoittotpsdass, chaudhuri2013refate, mehrabi2024opeimdpuwdo}, or adjusting on representations with better overlap, which we detail next.

\paragraph{Learning Representations as Adjustment Sets.} There is a substantial literature on adjusting for representations, i.e., deterministic mappings of covariates \citep{leacy2014otjuopapsieotateottass, lee2022racoteeupaps}. Such representations should preserve confounding information; extreme examples are balancing and prognostic scores, with exact definitions in Section \ref{sec:deconf_scores}. Those representations are generally not given and need to be estimated; classical approaches learn scalar representations using linear/logistic regression \citep{hansen2008tpaotps, leacy2014otjuopapsieotateottass} or focus on subsets of covariates \citep{schneeweiss2009hdpsaisoteuhccd} while more recent approaches leverage the compositional nature of neural networks to extract multivariate representations \citep{shalit2017eitegbaa, chernozhukov2022rafadmlwnnarf, clivio2022nsmfhdci}. Errors in estimation can lead to loss of confounding information; such loss is typically either not checked or assumed not to exist \citep{shalit2017eitegbaa, melnychuk2025orlfecq}, while more recent approaches attempt to quantify or minimize this confounding loss \citep{johansson2019saiidir, melnychuk2023boricbftee, clivio2024trlfwpidbci}. In contrast, \textit{we provide representations with no loss of confounding information}, generalizing balancing or prognostic scores. 

\paragraph{Learning Representations to Improve Overlap.} Researchers typically adjust for prognostic scores to reduce asymptotic variance \citep{austin2007acotaodpsmtbmvbtausamcs, schuler2021iteortevlafaps}; 
this approach has been referred to as \emph{collaborative} in the Targeted Machine Learning Estimation (TMLE) literature \citep{benkeser2020anseeotate, rudolph2023etateuassoem}. Typically, overlap in prognostic scores between treated and control groups is considered less stringent than overlap in original features \citep{luo2017oerbceusdr, damour2020oioswhdc, wu2021biviaeceulo}. 
In contrast, balancing scores are not used to improve overlap; instead, overlap is improved by removing non-confounding variables that predict the treatment assignment \citep{rubin1997ecefldsups, wooldridge2016sivbuamv}.
Other approaches, inspired by domain adaptation, minimize an objective that balances a treatment effect regression error against measures of poor overlap, including distributional distances \citep{shalit2017eitegbaa, johansson2022gbarlfeopoace}, support discrepancy measures \citep{johansson2019saiidir}, and conditional outcome posterior variances \citep{zhang2020lorfteoite}. However, to the best of our knowledge, none of these measures directly connects poor overlap to inferential challenges such as estimator variance. In contrast, our overlap divergence explicitly links the degree of overlap with respect to a representation to the asymptotic variance of estimators adjusting on that representation. Some of these approaches further incorporate inverse propensity weights in the objective, both in the regression error and in the measure of poor overlap \citep{assaad2021crlwbw, johansson2022gbarlfeopoace}. While this approach can improve treatment effect estimation, it is orthogonal to finding representations with better overlap. We further compare our approach to these lines of work in Appendix \ref{sec:comparison_with_domain_adaptation_approaches}.

\section{PRELIMINARIES}

\label{sec:preliminaries}
Let $(X_i, \binvar_i, Y_i) \stackrel{\text{i.i.d.}}{\sim} \Pall$ be i.i.d. samples of covariates $X$, a binary treatment $\binvar$, and an outcome $Y$. Denote $\Ps := \Pall(. | \binvar = 0)$, $\Pt := \Pall(. | \binvar = 1)$, $\pi_1 := \Pall(\binvar=1)$. We assume $0 < \pi_1 < 1$. Denote for any random variable $Z$ and distribution $R$, $R_Z$ the law of $Z$ in $R$, $\EP{R}$ the expectation wrt $R$, $m_0(Z) := \Es\left[Y|Z\right]$ and $m_1(Z) := \Et\left[Y|Z\right]$ the outcome models wrt $Z$, $\Delta m(Z) = m_1(Z) - m_0(Z)$ their difference, and $e(Z) := p(\binvar = 1 | Z)$ which is called the propensity score wrt $Z$. Note that, when relevant, the superscript of a distribution denotes the treatment indicator and its subscript is a random variable.

We focus on estimating the \emph{Average Treatment effect on the Treated} (ATT), $\tau = \Et[Y(1) - Y(0)]$, where $Y(1)$ and $Y(0)$ denote the potential outcomes under treatment and control, respectively.\footnote{Our results can be readily extended to general covariate shift, full population average treatment effect estimation or transportability \citep{clivio2024trlfwpidbci}.}
Throughout, we assume unconfoundedness, (one-sided) overlap wrt $X$, and technical assumptions about the outcome $Y$ and the density ratio between control and treatment distributions of $X$.
\begin{assumption}
(Unconfoundedness) All potential confounders of the relationship between treatment and potential outcomes are included in covariates $X$:
\begin{align*} 
(Y(0), Y(1)) \perp_{\Pall} T \mid X.
\end{align*}
\end{assumption}
\begin{assumption} \label{ass:overlap}
(One-sided overlap) $\Pt_X$ is absolutely continuous wrt $\Ps_X$, or equivalently, $e(X) < 1$ $\Pall$-almost surely.
\end{assumption}

\begin{assumption} \label{ass:strong_overlap}
    (Square-integrability of $\density{X}{\Ps}{\Pt}(X)$) The density ratio $\density{X}{\Ps}{\Pt}(X)$ between control and treatment distributions of $X$ is square-integrable in $P^0$:
    \begin{align*}
        \EP{\Ps}\left[\left(\density{X}{\Ps}{\Pt}(X)\right)^2\right]  < \infty.
    \end{align*}
\end{assumption}

\begin{assumption} \label{ass:y}
(Square-integrability of $Y$) The observed outcome $Y$ is square-integrable in $\Pall$:
\begin{align*}
\EP{\Pall}[Y^2] < \infty.
\end{align*}
\end{assumption}

Assumption \ref{ass:overlap} is a generalization of the standard overlap assumption, which states that every unit has some non-zero chance of receiving either treatment condition (or, equivalently, that there are no values of the covariates such that units with these values are either all in treatment or all in control). For this paper, we focus on the one-sided version of this assumption, where every unit has some non-zero probability of having been assigned to control, since our target estimand is the ATT that ignores units which cannot receive the treatment (in contrast, the standard overlap assumption applies when the target estimand is the standard Average Treatment Effect (ATE)). A stricter version of Assumption \ref{ass:overlap} is Assumption \ref{ass:strong_overlap}, which is the minimal assumption on overlap ensuring that many expectations in our derivations are well-defined. This is still weaker than the strict overlap assumption of \citet{damour2021oioswhdc} which uniformly bounds the density ratio. Assumption \ref{ass:y} is a technical assumption that prohibits pathological outcome distributions; for example, all bounded outcomes satisfy this assumption. Under these assumptions, $\tau$ can be identified by adjusting for $X$, that is, $\tau = \tau_X$ where, for any $Z$,
\begin{align}
\tau_Z := \Et\left[Y\right] - \Et\left[m_0(Z)\right]. \label{eq:gformula}
\end{align}
We will consider the properties of statistical estimands $\tau_Z$ that adjust for variables $Z$ other than $X$. The minimal possible asymptotic variance of regular and asymptotically linear (RAL) estimators of $\tau_Z$ from samples of $P(Z, \binvar, Y)$ is the \textit{semiparametric efficiency bound} $V_{\text{eff}}^Z$ \citep{tsiatis2006stamd} given as \citep{hahn1998otrotpsieseoate} 
\begin{align*}
V_{\text{eff}}^Z = &\E_{\Pall}\left[\frac{e(Z)\text{Var}_{\Pt}(Y|Z) + \left(\Delta m(Z) - \tau_{Z}\right)^2e(Z)}{\pi_1^2}\right]\\
&+ \E_{\Ps}\left[\frac{\text{Var}_{\Ps}(Y|Z)}{1-\pi_1}\left(\densityst{Z}(Z)\right)^2\right].
\end{align*}

We can see that $V_{\text{eff}}^Z$ depends on the magnitude of the density ratio $\densityst{Z}$; this magnitude describes the strength of overlap between the distributions of $Z$ in the treatment and control groups. When $\densityst{Z}$ takes large values, even under one-sided overlap wrt $Z$, RAL estimators of $\tau_Z$ have high asymptotic variance. Thus, our goal will be to build representations $\rep(X)$ that improve overlap compared to $X$ while ensuring $\tau = \tau_{\rep(X)}$.

\section{DECONFOUNDING SCORES AND OVERLAP DIVERGENCE}
\subsection{Deconfounding Scores}
\label{sec:deconf_scores}

We now introduce \emph{deconfounding scores}. These are defined as representations $\rep(X)$ such that $\tau$ can be identified by adjusting only for $\rep(X)$, that is $\tau = \tau_{\rep(X)}$. 
We can therefore view deconfounding scores as preserving the confounding information in $X$.

A key property is that deconfounding scores can be characterized as representations that introduce zero ``confounding bias'', expressible in terms of an observable conditional covariance (all proofs in Appendix \ref{app:proofs}).

\begin{lemma} \label{lem:cb}
    For any $\rep$, the confounding bias equals $$\tau_{\rep(X)}-\tau= \Es\left[\text{Cov}_{\Ps}\left(m_0(X)  ,  \densityst{X}(X)  \middle|  \rep(X)\right)\right].$$
\end{lemma}

Setting the conditional covariance in Lemma~\ref{lem:cb} to zero serves as a constraint that deconfounding scores must satisfy.
It is straightforward to verify that both the propensity score $e(X)$ (the density ratio is a measurable function of $e(X)$) and the control outcome model $m_0(X)$ are deconfounding scores. More generally, any $\rep(X)$ such that $m_0(X)$ is a measurable function of $\rep(X)$ (a \emph{prognostic score}, generalizing \citet{hansen2008tpaotps}) or $e(X)$ is a measurable function of $\rep(X)$ (a \emph{balancing score} as introduced in \citet{rosenbaum1983tcrotpsiosfce}) is a deconfounding score.
Crucially, the constraint also defines a continuum of representations between these two extremes, which we explore in this paper. We further discuss this result in Appendix \ref{app:discussion_cb}.

\subsection{Overlap Divergence}
While deconfounding scores all yield the same ATT estimand $\tau$, they can have different overlap properties.
Here, we introduce the \emph{overlap divergence}, which we use to measure the degree of overlap wrt a representation. As we will show shortly, it is closely connected to the semiparametric efficiency bound. The overlap divergence of a random variable $Z$ is defined as
\begin{align*}
   \misoverlapst(Z) := \EP{\Ps}\left[\left(\density{Z}{\Ps}{\Pt}(Z)\right)^2\right] = \chi^2(\Pt_Z || \Ps_Z) + 1,
\end{align*}
where $\chi^2(\Pt_Z || \Ps_Z) = \EP{\Ps}\left[\left(\density{Z}{\Ps}{\Pt}(Z) - 1\right)^2\right]$ is the $\chi^2$-divergence between $\Ps_Z$ and $\Pt_Z$. 
$\misoverlapst(\rep(X))$ quantifies the lack of overlap wrt $\rep(X)$, as it reflects the amplitude of $\densityst{\rep(X)}$. On the one hand, $\misoverlapst(\rep(X))$ is minimized (and equal to 1) if and only if $\Ps_{\rep(X)} = \Pt_{\rep(X)}$, which represents perfect overlap;
however, note that such representations $\rep(X)$ are typically not deconfounding scores, e.g., when $\rep(X)$ is constant. On the other hand, when the one-sided overlap assumption is not satisfied wrt $\rep(X)$, that is $\Pt_{\rep(X)}$ is not absolutely continuous wrt $\Ps_{\rep(X)}$, we have $\mathcal{O}(\rep(X)) = \infty$. One-sided overlap wrt $\rep(X)$ does not guarantee that $\mathcal{O}(\rep(X)) < \infty$; for example, $\mathcal{O}(\rep(X))$ can be infinite if $e(\rep(X))$ approaches 1. Note that Assumption \ref{ass:strong_overlap} can equivalently be written as $\misoverlapst(X) < \infty$.

We now justify our overlap divergence as controlling the semiparametric efficiency bound, as we show next.

\begin{lemma} \label{lem:misoverlap}
    For any representation $\rep(X)$, we have:
    \begin{enumerate}
        \item If $\text{Var}_{\Ps}(Y|X) \geq \sigma^2$ for some $\sigma > 0$, then $V_{\text{eff}}^{\rep(X)} \geq \frac{\sigma^2}{1-\pi_1} \cdot \misoverlapst(\rep(X))$.
        \item If $Y$ is bounded by some constant $Y_{\text{max}} > 0$ then $V_{\text{eff}}^{\rep(X)} \leq \frac{5 \cdot Y_{\text{max}}^2}{\pi_1} + \frac{Y_{\text{max}}^2}{1-\pi_1}\misoverlapst(\rep(X)).$
    \end{enumerate}
\end{lemma}

Together, these two bounds show that the overlap divergence is tightly linked to the efficiency bound: Item~1 implies that reducing the overlap divergence is \textit{necessary} for reducing the efficiency bound, while Item~2 implies that it may be \textit{sufficient}. We further discuss this result in Appendix \ref{app:discussion_misoverlap}.

 \section{OPTIMIZING THE OVERLAP DIVERGENCE}

We now aim to minimize $\misoverlapst(\rep(X))$ subject to the constraint that $\rep(X)$ is a deconfounding score. We first establish general properties in the nonparametric case, then show that a prognostic score solves this optimization problem in a Gaussian design setting.

 \subsection{Nonparametric Case: Representations Always Improve Overlap}

 We show that representations always lead to better overlap than original covariates and formalize the longstanding intuition that information predictive only of treatment should be excluded.
\begin{lemma} \label{lem:bse}
    The improvement of overlap divergence induced by a representation $\rep(X)$ equals
    \begin{align*}
        \misoverlapst(X) - \misoverlapst(\rep(X)) &= \Es\left[\text{Var}_{\Ps}\left(\densityst{X}(X) \ \middle| \ \rep(X)\right)\right]\\
        &\geq 0
    \end{align*}
    and it upper-bounds the absolute confounding bias as
    \begin{align*}
        |\tau_{\rep(X)} - \tau| \leq &\sqrt{\Es\left[\text{Var}_{\Ps}\left(m_0(X) \ \middle| \ \rep(X)\right)\right]}\\
        &\times \sqrt{\misoverlapst(X) - \misoverlapst(\rep(X))}.
    \end{align*}
\end{lemma}
 The first part of Lemma~\ref{lem:bse} shows that a representation \emph{always} improves overlap relative to the original covariates. Moreover, the improvement equals a measure of how poorly $\rep(X)$ predicts the density ratio $\densityst{X}(X)$---and thus the propensity score $e(X)$---termed the \textit{balancing score error} in \citet{clivio2024trlfwpidbci}.
 This confirms that variables predictive of treatment but not outcome should be excluded to reduce variance, which is a common intuition throughout the literature and has been highlighted in, e.g.,  \citet{rubin1997ecefldsups}, \citet{brookhart2006vsfpsm}, \citet{wooldridge2016sivbuamv} and \citet{colnet2024rwtrctfgfseavs}. To the best of our knowledge, this is the first mathematical proof of this intuition.

 Note that when $\rep(X)$ is not a balancing score, its propensity score $e(\rep(X))$ will differ from the original propensity score $e(X)$. This is desirable, however: as long as $\rep(X)$ is a deconfounding score, unbiased estimation is preserved, and Lemma~\ref{lem:bse} shows that the overlap divergence is strictly reduced.

 Finally, the upper bound in Lemma~\ref{lem:bse} suggests that a representation achieving both zero confounding bias and minimal overlap divergence should be a prognostic score. In the next section, we confirm this in a tractable analytical setting.

 \subsection{Gaussian Design: Prognostic Scores Optimize Overlap}
 \label{sec:linear_gaussian}

In this section, we explore the properties of deconfounding scores in a simple setting where we can obtain analytical expressions for a family of deconfounding scores.
Specifically, we consider a setting where covariates are standard centered Gaussian variables, the treatment assignment and outcome model are generalized linear models (GLMs), and representations are linear. We study two variants.
First, consider the case where covariates are Gaussian in $\Ps$.

\begin{assumption} \label{ass:lineargaussian1}
    $\Ps_X = \mathcal{N}(0,I_d)$, $m_0(x) = m(\alpha'x)$, $\densityst{X}(x) =  \frac{h(\beta'x)}{\E_{Z \sim \mathcal{N}(0,1)}[h(Z)]}$, $\rep(x) = \rep_\gamma(x) := \gamma'x$ for $\alpha, \beta, \gamma$ unit vectors in $\mathbb{R}^d$ with $\alpha'\beta \in (-1,1)$, $h, m$ real functions with $h \geq 0$ and $0 < \E_{Z \sim \mathcal{N}(0,1)}[h(Z)] < \infty$.
\end{assumption}

Second, we consider the case where covariates are Gaussian in $\Pall$. This variant is motivated by the fact that it is often impractical to make assumptions on the distribution of covariates in either treatment group and on the density ratio between the treatment groups --- instead preferring assumptions on covariates at the whole population level and on the propensity score.

\begin{assumption} \label{ass:lineargaussian2}
    $\Pall_X = \mathcal{N}(0,I_d)$, $m_0(x) = m(\alpha'x)$, $e(x) = h(\beta'x)$,  $\rep(x) = \rep_\gamma(x) := \gamma'x$ for $\alpha, \beta, \gamma$ unit vectors in $\mathbb{R}^d$ with $\alpha'\beta \in (-1,1)$, $h, m$ real functions with $0 \leq h(.) < 1$.
\end{assumption}

\subsubsection{Analytical Characterization of Deconfounding Scores}
We now show that a family of deconfounding scores can be computed in closed form in both settings.

\begin{theorem} \label{th:hyperbola}
If either Assumption \ref{ass:lineargaussian1} or Assumption \ref{ass:lineargaussian2} holds then $\rep_\gamma(X)$ is a deconfounding score for any $\gamma$ of the form $\gamma = w_1\frac{\alpha + \beta}{\sqrt{2 + 2\alpha'\beta}} + w_2\frac{\alpha - \beta}{\sqrt{2 - 2\alpha'\beta}} + n$ where $(1+\alpha'\beta)w_1^2 - (1-\alpha'\beta)w_2^2 = 2\alpha'\beta$, $w_1^2+w_2^2\leq1$, $n$ has norm $ \sqrt{1-w_1^2-w_2^2}$ and is in $\text{Null}(\text{Span}(\alpha, \beta))$. We refer to the set of such $\gamma$'s as $\mathcal{D}_{\alpha,\beta}$.
\end{theorem}

\begin{figure*}[t!]
\includegraphics[width=\textwidth]{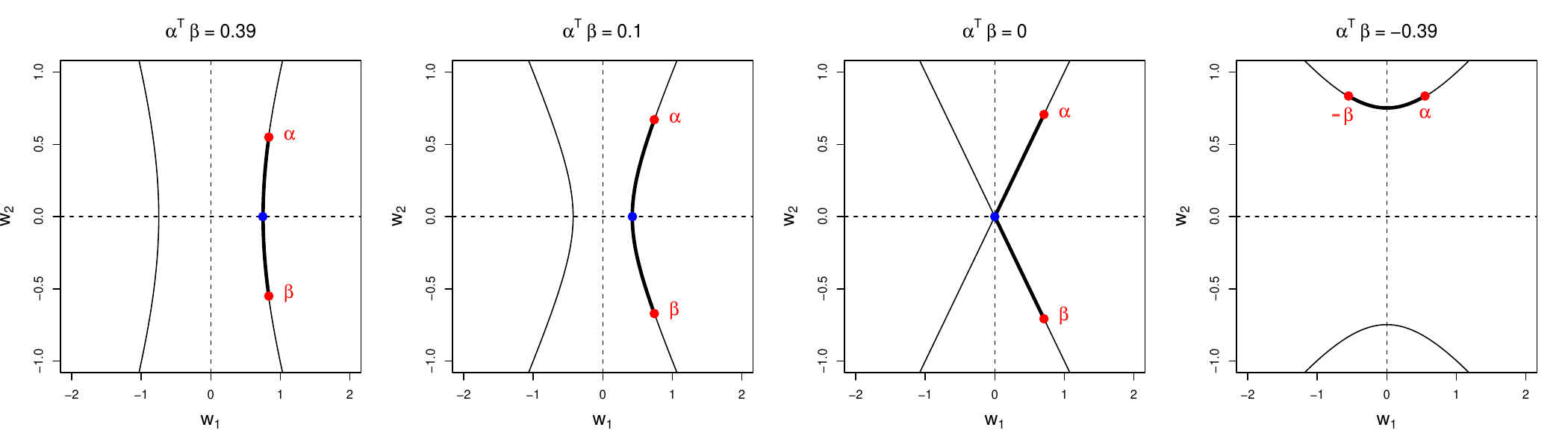}
\caption{The projection of $\gamma$ onto the space spanned by $\alpha$ and $\beta$ lies on a segment of a hyperbola (bold black line) whose endpoints correspond to $\gamma=\alpha$ and $\gamma=\beta$ (when $\alpha'\beta \geq 0$) or to $\gamma=\alpha$ and $\gamma=-\beta$ (when $\alpha'\beta < 0$), shown here, and the opposite segment, not shown here.  The orientation of the hyperbola and the endpoints depends on  $\alpha'\beta$. \label{fig:hyperbola}}
\end{figure*}

Note that $\mathcal{D}_{\alpha,\beta}$ does not depend on $h$ and $m$. To interpret this result, we note that the coordinates $(w_1, w_2)$ that yield valid unit-length values of $\gamma$ trace out two opposite segments of a hyperbola that lies on the subspace spanned by the prognostic score and propensity score coefficient vectors $\alpha$ and $\beta$. One segment has endpoints $\alpha$ and $\beta$ (when $\alpha'\beta \geq 0$) or $-\beta$ (when $\alpha'\beta < 0$), and the other has endpoints $-\alpha$ and $-\beta$ (when $\alpha'\beta \geq 0$) or $\beta$ (when $\alpha'\beta < 0$). Figure \ref{fig:hyperbola} shows the former segment for different values of $\alpha'\beta$.

When $\alpha'\beta \geq 0$, $w_2$ controls the position of the projection of $\gamma$ onto $\mathrm{Span}(\alpha, \beta)$ on either branch of the hyperbola. On the branch with endpoints $\alpha$ and $\beta$, when we move $w_2$ along its valid range $\left[-\sqrt{\frac{1-\alpha'\beta}{2}}, \sqrt{\frac{1-\alpha'\beta}{2}}\right]$, setting $w_2$ to its maximal value implies that $\gamma = \alpha$ so that $\rep_\gamma(X)$ is a prognostic score, whereas setting $w_2$ to its minimal value implies that $\gamma = \beta$ so that $\rep_\gamma(X)$ is a balancing score.  $w_2 = 0$ implies that $\gamma$ is equiangular to $\alpha$ and $\beta$, that is $\alpha'\gamma = \beta'\gamma$. An analogous description can be made for the branch with endpoints $-\alpha$ and $-\beta$.
For points on the interior of $w_2$'s range, there is an equivalence class of $\gamma$'s: the projection of $\gamma$ onto $\mathrm{Span}(\alpha, \beta)$ is strongly constrained, but the orthogonal component of $\gamma$ is only constrained to make sure that $\gamma$ has norm 1. We interpret $w_2$ as a scalar parameter that controls the similarity of the deconfounding score to the balancing and prognostic scores.

Note that $w_1$ can be analogously interpreted when $\alpha'\beta \leq 0$. In this case, the segment with prognostic score endpoint $\alpha$ has $-\beta$ as a propensity score endpoint; the other segment has prognostic score and propensity score endpoints $-\alpha$ and $\beta$, respectively. Equivalently, we can enforce $\alpha'\beta > 0$ and replace $\beta$ with $-\beta$ when $\alpha'\beta < 0$; we justify this in Appendix \ref{app:discussion_hyperbola}.

\subsubsection{Optimality of Prognostic Scores}

Given this family of deconfounding scores, we might suspect from Lemma~\ref{lem:bse} that its prognostic scores, which have no explicit dependence on the propensity score or the density ratio, would be overlap-optimal.
Here, we show that this is in fact the case.

\begin{theorem} \label{th:misoverlap_prognostic_score}
    Defining, for any integer $K$ and for fixed $C > 1$, $0 \leq \lambda < \frac{1}{4}$, $0 \leq \lambda' < \frac{1 - 4\lambda}{6}$,
    \begin{align*}
        \mathcal{H}^K_{C,\lambda} &:= \{h \ | \ h \text{ is $K$ times differentiable, }  \\
        & \ \ \ \ \ \ \ \ \ \ \forall k = 0, .., K, \ \forall z, \ |h^{(k)}(z)| \leq Ce^{\lambda z^2} \}, \\
        \mathcal{H}'^K_{C,\lambda,\lambda'} &:= \{h \ | \ h \text{ is $K$ times differentiable, }\\
        & \ \ \ \ \ \ \ \ \ \   \forall z, \  1 - h(z) \geq \frac{e^{-\lambda' z^2}}{C}, \ h(z) \geq 0,\\
        & \ \ \ \ \ \ \ \ \ \  \forall k = 1, .., K, \  \forall z, \ |h^{(k)}(z)| \leq Ce^{\lambda  z^2} \},
    \end{align*}
    we have:
    \begin{enumerate}
        \item If either (a) Assumption \ref{ass:lineargaussian1} holds with $h \in \mathcal{H}^2_{C,\lambda}$, or (b) Assumption \ref{ass:lineargaussian2} holds with $h \in \mathcal{H}'^2_{C,\lambda,\lambda'}$, then
            \begin{enumerate}[label=(\Alph*)]%
            \item $\misoverlapst(\rep_\gamma(X))$ is non-decreasing in $|\beta'\gamma|$.
            \item If $\alpha'\beta \neq 0$, $\misoverlapst(\rep_\gamma(X))$ is non-decreasing when moving from $\alpha$ to $\beta$ (when $\alpha'\beta > 0$) or to $-\beta$ (when  $\alpha'\beta < 0$), and when moving from $-\alpha$ to $-\beta$  (when $\alpha'\beta > 0$) or to $\beta$ (when  $\alpha'\beta < 0$), on the corresponding portion of $\mathcal{D}_{\alpha,\beta}$; notably, $\gamma = \alpha$ and $\gamma = -\alpha$ are global minimizers of $\misoverlapst(\rep_\gamma(X))$ on $\mathcal{D}_{\alpha,\beta}$.
            \item If $\alpha'\beta = 0$,  the $\gamma$'s whose projection onto $\mathrm{Span}(\alpha, \beta)$ belongs to $[-\alpha,\alpha]$ are global optimizers of $\misoverlapst(\rep_\gamma(X))$ on $\mathcal{D}_{\alpha,\beta}$.
            \end{enumerate}
        \item If either (a) Assumption \ref{ass:lineargaussian1} holds with (i) $h \in \mathcal{H}^{K+1}_{C,\lambda}$ for some integer $K \geq 2$ such that  $\E_{Z \sim \mathcal{N}(0,1)}\left[h^{(K)}(Z)\right] \neq 0$, or (ii) $h(z) = 1_{\{z \leq z_0\}}$ for some $z_0 \in \mathbb{R}$, or (iii) $h(z) = \text{ReLU}(z)$, or if (b) Assumption \ref{ass:lineargaussian2} holds with $h \in \mathcal{H}'^{K+1}_{C,\lambda,\lambda'}$ for some integer $K \geq 2$ such that  $\E_{Z \sim \mathcal{N}(0,1)}\left[h^{(K)}(Z)\right] \neq 0$, then we obtain the same (A), (B) and (C) as in 1. but where ``non-decreasing'' is replaced with ``increasing'' and ``global minimizers'' with ``the only global minimizers''.
    \end{enumerate}
\end{theorem}

Under the assumptions of Theorem \ref{th:misoverlap_prognostic_score}, the overlap divergence of $\rep_\gamma(X)$ is non-decreasing (or strictly increasing) in $|\beta'\gamma|$, which measures the strength of the association between the treatment assignment (parameterized by $\beta$) and the representation (parameterized by $\gamma$), consistent with Lemma \ref{lem:bse}. In the generic case ($\alpha'\beta \neq 0$), overlap improves monotonically as we move along either segment of the hyperbola from the balancing score toward the prognostic score, strengthening Lemma~\ref{lem:bse}. In the degenerate case ($\alpha'\beta = 0$), half of either segment is optimal, but the prognostic scores always remain among the optimizers. We emphasize the key conclusion: \textit{among deconfounding scores in $\mathcal{D}_{\alpha, \beta}$, the prognostic scores yield the best overlap.}
 We further discuss assumptions in Appendix \ref{app:discussion_misoverlap_prognostic_score}.

\section{EXPERIMENTS}
\label{sec:results}
We now assess how our analytical results translate to finite-sample estimation. We evaluate ATT analogues of the outcome regression \citep{hahn1998otrotpsieseoate}, IPW \citep{horvitz1952agoswrfafu}, and AIPW \citep{robins1994eorcwsranao} estimators, replacing covariates $X$ with linear deconfounding scores $\rep_\gamma(X)$ expressed as in the $\mathcal{D}_{\alpha, \beta}$ set of Theorem \ref{th:hyperbola}. We give further details in Appendix \ref{app:xps}. The code to reproduce experiments is available at \url{https://github.com/oscarclivio/deconfounding_scores_paper}.

\subsection{Estimators and Inference}
\label{sec:estimators_and_inference}

We consider canonical ATT estimators: (i) the outcome regression estimator, obtained by plugging the estimated outcome model into Equation \ref{eq:gformula} (``Regr''), (ii) the IPW estimator (``IPW''), and (iii) the AIPW estimator (``AIPW''). We compare these to the analogous estimators where input features are deconfounding scores $\rep_\gamma(X)$. They are denoted as ``Method-$\gamma$'', where ``Method'' refers to the base methods ``Regr'', ``IPW'', ``AIPW'' and $\gamma$ is the coefficient vector of the deconfounding score passed to the base method. We estimate the outcome and propensity models using either LASSO regression or Ridge regression. Regularization parameters for models fit to original features are selected via cross-validation using the \texttt{glmnet} R package \citep{friedman2010rpfglmvcd, rcoreteam2024r}. We do not use regularization when estimating models wrt one-dimensional deconfounding scores.

To estimate deconfounding scores, we use coefficient vectors $\hat{\alpha} = \hat{\alpha}^1$ and $\hat{\beta} = \text{sign}(\hat{\alpha}^{1\prime}\hat{\beta}^1)\hat{\beta}^1$ where $\hat{\alpha}^1$ and $\hat{\beta}^1$ are the normalized coefficient vectors obtained through the above LASSO or Ridge regression with respect to original features. To ensure $\hat{\alpha}'\hat{\beta} \geq 0$, we replace $\hat{\beta}^1$ with $-\hat{\beta}^1$ whenever $\hat{\alpha}^{1\prime}\hat{\beta}^1 < 0$. These are used as plug-ins in the set $\mathcal{D}_{\hat{\alpha} , \hat{\beta}}$ given in Theorem~\ref{th:hyperbola} to yield estimated linear deconfounding scores $\rep_{\hat{\gamma}}(X)$; we sample one orthogonal component $\hat{n}$ with appropriate normalization in $\text{Null}(\text{Span}(\hat{\alpha}, \hat{\beta}))$ using the \texttt{rstiefel} R package \citep{hoff2013bayesiananalysismatrixdata}. We parameterize  $\mathcal{D}_{\hat{\alpha},  \hat{\beta}}$ using  a normalized version $w$ of $w_2$, where $w_2 = -\sqrt{\frac{1-\hat{\alpha}'\hat{\beta}}{2}} \times w$. Here, (i) $w=1$ indicates that $\rep_{\hat{\gamma}}(X) = \hat{\beta}'X$, which is an estimated balancing score; (ii) $w=-1$ means $\rep_{\hat{\gamma}}(X) = \hat{\alpha}'X$, which is an estimated prognostic score; (iii) $w=0$ gives a $\rep_{\hat{\gamma}}(X)$ that is equiangular to the estimated prognostic and balancing scores on the hyperbola; we denote its coefficient vector by $\hat{\delta}$ and its ground-truth analogue by $\delta$.

\subsection{Results on a Simulated Dataset}
Using the model given in Assumption \ref{ass:lineargaussian2}, we set $m$ to the identity and $h$ to the inverse logit function.
We generate $n$ i.i.d.\ triples $(X_i, \binvar_i, Y_i)$ according to
\begin{align*}
X &\sim \mathcal{N}(0, I_p),\\
T &\sim \text{Bernoulli}(e(X)), \ \ e(X) = \text{logit}^{-1}(\beta_0 + s_TX'\beta),\\
Y  &\sim \mathcal{N}(\alpha_0 + s_YX'\alpha + \tau T, 1).
\end{align*}

Here, $s_Y$ corresponds to the signal-to-noise ratio (SNR) for the outcome model, while $s_T$ controls overlap; higher values of $s_T$ correspond to poorer overlap. $\alpha$ and $\beta$ are constructed to share the same 20-element support with  $\alpha'\beta = 0.75$. 
We take $n=500$, $p=1000$ and $\tau = 0$.  We consider high overlap $s_T = 1$ and low overlap $s_T = 4$,  as well as  $s_Y=2$ and $s_Y=5$ with larger values implying a higher SNR. 
Since the true outcome and propensity models are sparse, LASSO with appropriate variable selection is correctly specified, while Ridge, which does not perform variable selection, is misspecified. We report averages across 100 runs.

\begin{table}[h!]
\caption{RMSEs on simulated datasets.
\green{Green}: $\rep_{\hat{\gamma}}(X)$ \green{improves over} $X$ for the same base method; \red{Red}: it is \red{worse}. Best performance \underline{underlined}. See Section \ref{sec:estimators_and_inference} for definitions of names of estimators.
}
\vspace{0.3cm}
\label{tab:rmse} 
\centering
\resizebox{!}{0.42\textheight}{
\begin{tabular}{l|c|c|c|c}

\multicolumn{1}{c|}{Overlap} & \multicolumn{2}{c|}{High} & \multicolumn{2}{c}{Low} \\
\cline{1-1} \cline{2-3} \cline{4-5}
\multicolumn{1}{c|}{SNR } & \multicolumn{1}{c|}{Low} & \multicolumn{1}{c|}{High} & \multicolumn{1}{c|}{Low} & \multicolumn{1}{c}{High} \\
\cline{1-1} \cline{2-2} \cline{3-3} \cline{4-4} \cline{5-5}
\hline
\multicolumn{5}{c}{ LASSO outcome and propensity } \\

\hline
IPW & 0.98 & 2.45 & 1.45 & 3.59\\

AIPW & 0.35 & 0.39 & 0.79 & 1.04\\

Regr & 0.4 & 0.43 & 1 & 1.26\\

\hline
IPW-$\hat{\beta}$ & \green{0.66} & \green{1.58} & \green{1.1} & \green{2.46}\\

AIPW-$\hat{\beta}$ & \red{0.59} & \red{1.39} & \red{1.13} & \red{2.6}\\

Regr-$\hat{\beta}$ & \red{0.47} & \red{1.12} & \green{0.68} & \red{1.62}\\

\hline
IPW-$\hat{\delta}$ & \green{0.43} & \green{1.01} & \green{0.5} & \green{0.96}\\

AIPW-$\hat{\delta}$ & \red{0.39} & \red{0.88} & \green{0.37} & \green{0.65}\\

Regr-$\hat{\delta}$ & \green{0.39} & \red{0.88} & \underline{\green{0.31}} & \underline{\green{0.46}}\\

\hline
IPW-$\hat{\alpha}$ & \green{0.27} & \green{0.56} & \green{0.55} & \green{0.96}\\

AIPW-$\hat{\alpha}$ & \underline{\green{0.22}} & \green{0.26} & \green{0.65} & \green{0.9}\\

Regr-$\hat{\alpha}$ & \green{0.22} & \underline{\green{0.25}} & \green{0.64} & \green{0.89}\\

\hline
\multicolumn{5}{c}{ LASSO outcome and Ridge propensity } \\

\hline
IPW & 1.17 & 2.93 & 2.1 & 5.24\\

AIPW & 0.39 & 0.42 & 0.98 & 1.23\\

Regr & 0.4 & 0.43 & 1 & 1.26\\

\hline
IPW-$\hat{\beta}$ & \green{1.14} & \green{2.7} & \green{1.92} & \green{4.57}\\

AIPW-$\hat{\beta}$ & \red{1.15} & \red{2.71} & \red{1.51} & \red{3.44}\\

Regr-$\hat{\beta}$ & \red{0.47} & \red{1.09} & \green{0.84} & \red{1.99}\\

\hline
IPW-$\hat{\delta}$ & \green{0.32} & \green{0.71} & \green{0.79} & \green{1.96}\\

AIPW-$\hat{\delta}$ & \green{0.28} & \red{0.61} & \green{0.74} & \red{1.59}\\

Regr-$\hat{\delta}$ & \green{0.22} & \red{0.45} & \green{0.57} & \red{1.26}\\

\hline
IPW-$\hat{\alpha}$ & \underline{\green{0.18}} & \green{0.29} & \underline{\green{0.55}} & \green{0.96}\\

AIPW-$\hat{\alpha}$ & \green{0.22} & \underline{\green{0.25}} & \green{0.65} & \green{0.9}\\

Regr-$\hat{\alpha}$ & \green{0.22} & \green{0.25} & \green{0.64} & \underline{\green{0.89}}\\

\hline
\multicolumn{5}{c}{ Ridge outcome and LASSO propensity } \\

\hline
IPW & 0.98 & 2.45 & 1.45 & 3.59\\

AIPW & 0.99 & 2.47 & 1.66 & 4.13\\

Regr & 1.12 & 2.81 & 2.06 & 5.16\\

\hline
IPW-$\hat{\beta}$ & \green{0.66} & \green{1.58} & \green{1.1} & \green{2.46}\\

AIPW-$\hat{\beta}$ & \green{0.61} & \green{1.45} & \green{1.13} & \green{2.6}\\

Regr-$\hat{\beta}$ & \underline{\green{0.5}} & \underline{\green{1.2}} & \underline{\green{0.68}} & \underline{\green{1.62}}\\

\hline
IPW-$\hat{\delta}$ & \green{0.66} & \green{1.64} & \green{1.29} & \green{3.27}\\

AIPW-$\hat{\delta}$ & \green{0.75} & \green{1.9} & \green{1.45} & \green{3.69}\\

Regr-$\hat{\delta}$ & \green{0.75} & \green{1.9} & \green{1.43} & \green{3.65}\\

\hline
IPW-$\hat{\alpha}$ & \green{0.89} & \green{2.22} & \red{1.79} & \red{4.46}\\

AIPW-$\hat{\alpha}$ & \red{1.02} & \red{2.55} & \red{1.94} & \red{4.86}\\

Regr-$\hat{\alpha}$ & \green{1.02} & \green{2.55} & \green{1.94} & \green{4.86}\\

\hline
\multicolumn{5}{c}{ Ridge outcome and propensity } \\

\hline
IPW & 1.17 & 2.93 & 2.1 & 5.24\\

AIPW & 1.09 & 2.73 & 2.02 & 5.04\\

Regr & 1.12 & 2.81 & 2.06 & 5.16\\

\hline
IPW-$\hat{\beta}$ & \green{1.14} & \green{2.7} & \green{1.92} & \green{4.57}\\

AIPW-$\hat{\beta}$ & \red{1.15} & \green{2.71} & \green{1.51} & \green{3.44}\\

Regr-$\hat{\beta}$ & \underline{\green{0.47}} & \underline{\green{1.09}} & \underline{\green{0.84}} & \underline{\green{1.99}}\\

\hline
IPW-$\hat{\delta}$ & \green{0.51} & \green{1.22} & \green{1.37} & \green{3.38}\\

AIPW-$\hat{\delta}$ & \green{0.55} & \green{1.37} & \green{1.46} & \green{3.62}\\

Regr-$\hat{\delta}$ & \green{0.55} & \green{1.35} & \green{1.43} & \green{3.58}\\

\hline
IPW-$\hat{\alpha}$ & \green{0.88} & \green{2.19} & \green{1.79} & \green{4.46}\\

AIPW-$\hat{\alpha}$ & \green{1.01} & \green{2.54} & \green{1.94} & \green{4.86}\\

Regr-$\hat{\alpha}$ & \green{1.01} & \green{2.54} & \green{1.94} & \green{4.86}\\

\end{tabular}

}

\end{table}

\begin{figure}[h!]
\centering
    \centering 
        \includegraphics[width=\columnwidth]{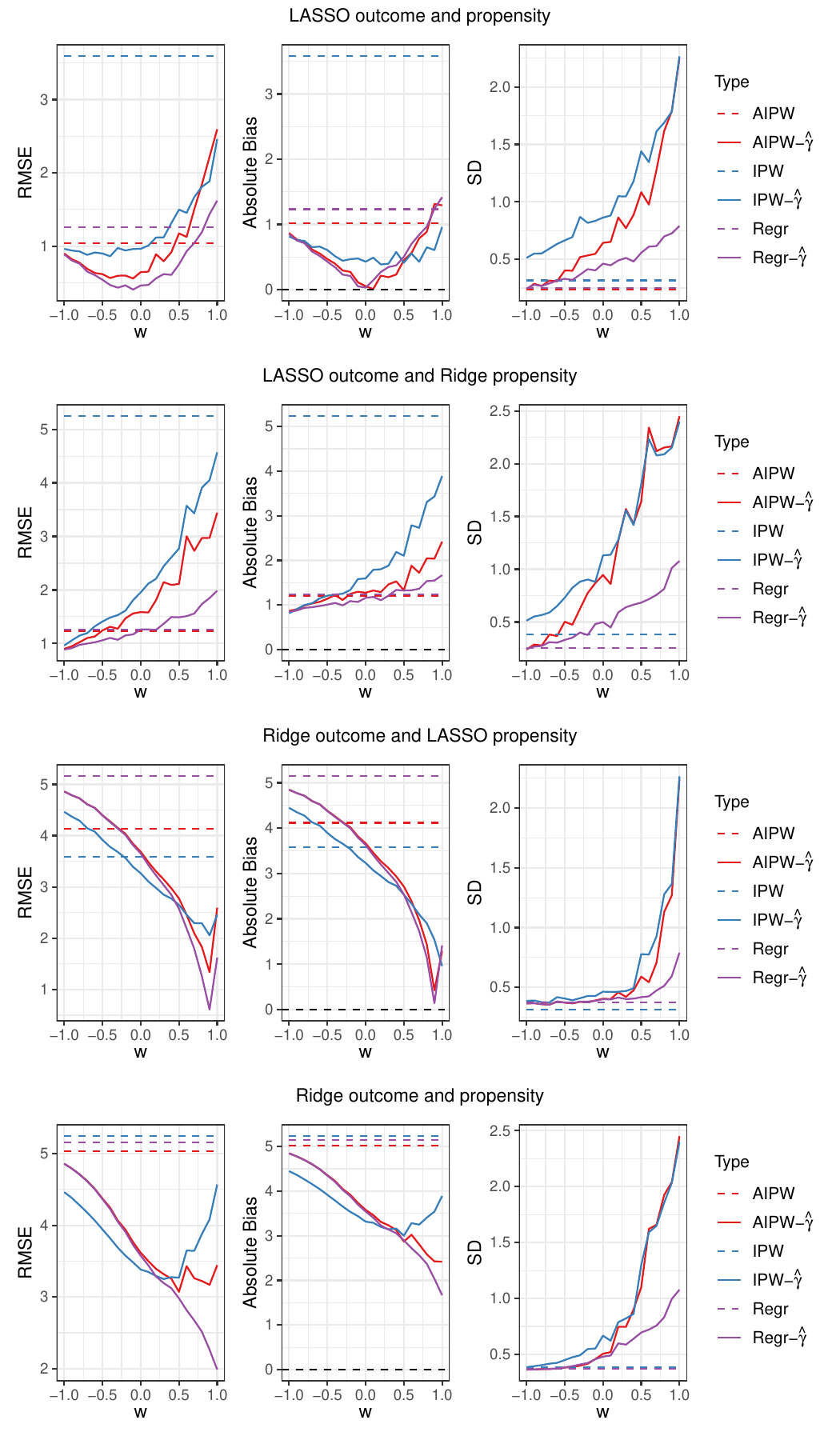}
        \label{fig:bv_est}
\caption{\textbf{RMSE, bias and standard deviation for simulated datasets.} Each metric is plotted according to the deconfounding score coordinate parameter $w$; methods using base covariates are constant. In all plots, $s_T = 4$ (low overlap) and $s_Y = 5$ (high SNR). See Section \ref{sec:estimators_and_inference} for definitions of names of estimators. 
\label{fig:bias_var}}
\end{figure}

\paragraph{Results on the Overall RMSE.} Table \ref{tab:rmse} reports the root mean squared error (RMSE) of ATT estimates using either the original covariates or three deconfounding scores: the estimated balancing score, prognostic score, and equiangular score. Several patterns emerge. When the outcome model is well-specified, prognostic-score-based estimators always improve over original estimators and often have the lowest RMSE, particularly when the propensity model is misspecified. When the outcome model is misspecified but the propensity model is well-specified, the pattern reverses: balancing-score-based methods dominate. When both models are misspecified, deconfounding scores nearly always outperform raw covariates. Overall, the equiangular score is an attractive middle ground: it benefits from combining information from both models and avoids overfitting to a single poorly-specified score. We detail this next.

\paragraph{Decomposition into Bias and Variance.} To explain the results in Table \ref{tab:rmse}, Figure~\ref{fig:bias_var} plots the RMSE, absolute bias, and standard deviation (SD) of estimators for $s_T=4$ and $s_Y=5$ across deconfounding scores parameterized by $w \in \{-1, -0.9, \dots, 0.9, 1\}$. The SD generally decreases as $w$ moves from $1$ (estimated balancing score) toward $-1$ (estimated prognostic score), consistent with the theory, although deconfounding-score-based estimators have somewhat higher SDs than those using raw covariates. Since RMSE is dominated by bias in these experiments, the SD patterns are less consequential for overall performance.

The bias patterns are more informative and track model specification closely: (i)~when both outcome and propensity models are well-specified, intermediate deconfounding scores exhibit lower bias, likely because they combine information from both models; (ii)~when only one model is well-specified, bias is lowest near the corresponding endpoint---the prognostic score when the outcome model is correct, the balancing score when the propensity model is correct---and increases toward the misspecified endpoint; (iii)~when both models are misspecified, bias for deconfounding scores is generally lower than for raw covariates.

\subsection{Results on Semi-Synthetic Datasets}

We now assess these estimators on canonical semi-synthetic datasets: IHDP \citep{hill2011bnmfci}, ACIC 2016 \citep{dorie2017avdiymfcillfadac} and HC-MNIST \citep{jesson2021qiiilceeuhc}. Critically, they do not satisfy Assumptions \ref{ass:lineargaussian1} and \ref{ass:lineargaussian2}. IHDP offers 6 different settings to generate the data, ACIC 2016 offers 77 settings, and HC-MNIST offers 1 setting. Thus, we report RMSEs over both settings and runs for each of these datasets. We conduct 100 runs. Results are in Table \ref{tab:rmse_semisynth}.  We note that the best performance is always achieved by a deconfounding score. Performance of each individual type of deconfounding score depends on the dataset and specification of models. Overall, estimated prognostic scores tend to outperform original covariates more frequently than other scores, consistent with the intuition from the theory. However, other scores can offer superior performance depending on the dataset, notably the equiangular score which again proves to be an attractive alternative. We hypothesize that this overall behavior depends on which model (outcome or propensity) is adequately captured by the corresponding learned coefficient vector ($\hat{\alpha}$ or $\hat{\beta}$, respectively) under misspecification, with the equiangular score potentially offering a robust compromise. 

\begin{table}[h!]
\caption{RMSEs on semi-synthetic datasets.
\green{Green}: $\rep_{\hat{\gamma}}(X)$ \green{improves over} $X$ for the same base method; \red{Red}: it is \red{worse}. Best performance \underline{underlined}. See Section \ref{sec:estimators_and_inference} for definitions of names of estimators.
}
\vspace{0.3cm}
\label{tab:rmse_semisynth} 
\centering
\resizebox{!}{0.42\textheight}{
\begin{tabular}{l|c|c|c}

\multicolumn{1}{c|}{Dataset } & \multicolumn{1}{c|}{IHDP} & \multicolumn{1}{c|}{ACIC2016} & \multicolumn{1}{c}{HC-MNIST} \\
\cline{1-1} \cline{2-2} \cline{3-3} \cline{4-4}
\hline
\multicolumn{4}{c}{ LASSO outcome and propensity } \\

\hline
IPW & 2.35 & 2.29 & 0.2\\

AIPW & 2.41 & 2.09 & 0.22\\

Regr & 2.41 & 0.62 & 0.22\\

\hline
IPW-$\hat{\beta}$ & \red{2.46} & \red{2.48} & \green{0.18}\\

AIPW-$\hat{\beta}$ & \red{2.43} & \red{2.41} & \red{0.22}\\

Regr-$\hat{\beta}$ & \red{2.46} & \red{0.64} & \underline{\green{0.14}}\\

\hline
IPW-$\hat{\delta}$ & \underline{\green{2.21}} & \green{1.27} & \red{0.49}\\

AIPW-$\hat{\delta}$ & \green{2.21} & \green{1.41} & \red{0.96}\\

Regr-$\hat{\delta}$ & \green{2.21} & \red{0.97} & \red{0.5}\\

\hline
IPW-$\hat{\alpha}$ & \red{2.44} & \green{1.04} & \green{0.17}\\

AIPW-$\hat{\alpha}$ & \red{2.43} & \green{0.72} & \green{0.21}\\

Regr-$\hat{\alpha}$ & \red{2.41} & \underline{\green{0.6}} & \green{0.2}\\

\hline
\multicolumn{4}{c}{ LASSO outcome and Ridge propensity } \\

\hline
IPW & 2.38 & 1.78 & 0.21\\

AIPW & 2.41 & 1.61 & 0.22\\

Regr & 2.41 & 0.62 & 0.22\\

\hline
IPW-$\hat{\beta}$ & \red{2.53} & \red{2.24} & \green{0.19}\\

AIPW-$\hat{\beta}$ & \red{2.48} & \red{2.16} & \red{0.22}\\

Regr-$\hat{\beta}$ & \red{2.53} & \red{0.63} & \underline{\green{0.14}}\\

\hline
IPW-$\hat{\delta}$ & \underline{\green{2.24}} & \green{1.08} & \red{0.48}\\

AIPW-$\hat{\delta}$ & \green{2.24} & \green{1.21} & \red{0.48}\\

Regr-$\hat{\delta}$ & \green{2.24} & \red{1.03} & \red{0.49}\\

\hline
IPW-$\hat{\alpha}$ & \red{2.44} & \green{1.04} & \green{0.17}\\

AIPW-$\hat{\alpha}$ & \red{2.43} & \green{0.72} & \green{0.21}\\

Regr-$\hat{\alpha}$ & \red{2.41} & \underline{\green{0.6}} & \green{0.2}\\

\hline
\multicolumn{4}{c}{ Ridge outcome and LASSO propensity } \\

\hline
IPW & 2.35 & 2.29 & 0.2\\

AIPW & 2.4 & 2.04 & 0.22\\

Regr & 2.4 & 0.62 & 0.27\\

\hline
IPW-$\hat{\beta}$ & \red{2.46} & \red{2.48} & \green{0.18}\\

AIPW-$\hat{\beta}$ & \red{2.43} & \red{2.41} & \red{0.22}\\

Regr-$\hat{\beta}$ & \red{2.46} & \red{0.64} & \underline{\green{0.14}}\\

\hline
IPW-$\hat{\delta}$ & \underline{\green{2.19}} & \green{1.3} & \red{0.67}\\

AIPW-$\hat{\delta}$ & \green{2.2} & \green{1.36} & \red{3.55}\\

Regr-$\hat{\delta}$ & \green{2.2} & \red{1.02} & \red{0.5}\\

\hline
IPW-$\hat{\alpha}$ & \red{2.45} & \green{1.02} & \green{0.19}\\

AIPW-$\hat{\alpha}$ & \red{2.43} & \green{0.69} & \red{0.23}\\

Regr-$\hat{\alpha}$ & \red{2.42} & \underline{\green{0.6}} & \green{0.22}\\

\hline
\multicolumn{4}{c}{ Ridge outcome and propensity } \\

\hline
IPW & 2.38 & 1.78 & 0.21\\

AIPW & 2.41 & 1.57 & 0.23\\

Regr & 2.4 & 0.62 & 0.27\\

\hline
IPW-$\hat{\beta}$ & \red{2.53} & \red{2.24} & \green{0.19}\\

AIPW-$\hat{\beta}$ & \red{2.48} & \red{2.16} & \green{0.22}\\

Regr-$\hat{\beta}$ & \red{2.53} & \red{0.63} & \underline{\green{0.14}}\\

\hline
IPW-$\hat{\delta}$ & \underline{\green{2.23}} & \green{1.12} & \red{0.61}\\

AIPW-$\hat{\delta}$ & \green{2.23} & \green{1.25} & \red{5.66}\\

Regr-$\hat{\delta}$ & \green{2.23} & \red{1.08} & \red{0.52}\\

\hline
IPW-$\hat{\alpha}$ & \red{2.45} & \green{1.02} & \green{0.19}\\

AIPW-$\hat{\alpha}$ & \red{2.43} & \green{0.69} & \red{0.23}\\

Regr-$\hat{\alpha}$ & \red{2.42} & \underline{\green{0.6}} & \green{0.22}\\

\end{tabular}

}
\end{table}

\section{CONCLUSION}

We have introduced deconfounding scores---representations subject to a zero-confounding-bias constraint---and shown that prognostic scores are overlap-optimal---as measured by overlap divergence--- within a family of deconfounding scores under Gaussian covariates and generalized linear models.
In experiments, there is always at least one deconfounding score improving ATT estimation over raw covariates, with simulations suggesting that performance of deconfounding scores is determined by the correctness of the estimated outcome and propensity models. Importantly, our approach is complementary to existing estimators: deconfounding scores can serve as drop-in replacements for covariates in any treatment effect estimation method, including AIPW \citep{robins1994eorcwsranao}, TMLE \citep{vanderlaan2011tlcifoaed}, and double/debiased machine learning \citep{chernozhukov2018ddmlftasp, chernozhukov2021admlvrr}.

Two key limitations merit future work. First, our analytical results rely on restrictive assumptions (Gaussian covariates, generalized linear models), and performance depends on correct specification of the outcome and propensity models; analyzing the impact of estimation error in the coefficient vectors is a natural next step.
Second, estimating overlap-optimal representations from finite samples in more general settings remains an open problem. One direction would be to generalize classical $\chi^2$-divergence optimization methods \citep{nguyen2010edfatlrbcrm, dieng2017vicubma, huggins2020vvivppeb} to non-trivial push-forward measures with a deconfounding score constraint. Additionally, while Lemma~\ref{lem:misoverlap} identifies the squared density ratio as the dominant term in the efficiency bound, other terms may become important when the density ratio is moderate; jointly optimizing the full efficiency bound with respect to the representation is a challenging but valuable direction. We discuss possible extensions more broadly in Appendix \ref{app:extensions}.

\newpage\phantom{...}

\newpage

\subsubsection*{Acknowledgements}

We sincerely thank anonymous reviewers for valuable feedback. O.C. was supported by Novo Nordisk and the U.K. Engineering and Physical Sciences Research Council through the Centre for Doctoral Training in Modern Statistics and Statistical Machine Learning (Project EP/S023151/1). A.D. is an employee of Google and may own stock as a part of a standard compensation package. Al.F. was supported in part by the U.S. National Institutes of Health through Grants 1R01GM144967-01 and 1R03CA211160-01, by the U.S. National Science Foundation through Award 1924205, and by the Chan Zuckerberg Initiative. D.B.-S. and Av.F. were supported in part by the Institute of Education Sciences, U.S. Department of Education, through Grants R305D200010 and R305D240036, and by the U.S. National Science Foundation through Award 2243822. C.H. was supported by the Alan Turing Institute, the Li Ka Shing Foundation, the Ellison Institute of Technology, the U.K. Engineering and Physical Sciences Research Council through the Bayes4Health grant EP/R018561/1, and U.K. Research and Innovation through the Medical Research Council and the ``AI and data science for engineering, health and government (ASG)'' programme.

\bibliographystyle{plainnat}
\bibliography{refs}

\section*{Checklist}

\begin{enumerate}

  \item For all models and algorithms presented, check if you include:
  \begin{enumerate}
    \item A clear description of the mathematical setting, assumptions, algorithm, and/or model. Yes
    \item An analysis of the properties and complexity (time, space, sample size) of any algorithm. Not Applicable
    \item (Optional) Anonymized source code, with specification of all dependencies, including external libraries. Yes
  \end{enumerate}

  \item For any theoretical claim, check if you include:
  \begin{enumerate}
    \item Statements of the full set of assumptions of all theoretical results. Yes
    \item Complete proofs of all theoretical results. Yes
    \item Clear explanations of any assumptions. Yes   
  \end{enumerate}

  \item For all figures and tables that present empirical results, check if you include:
  \begin{enumerate}
    \item The code, data, and instructions needed to reproduce the main experimental results (either in the supplemental material or as a URL). Yes
    \item All the training details (e.g., data splits, hyperparameters, how they were chosen). Yes
    \item A clear definition of the specific measure or statistics and error bars (e.g., with respect to the random seed after running experiments multiple times). No: measures of uncertainty were not included due to a lack of space.
    \item A description of the computing infrastructure used. (e.g., type of GPUs, internal cluster, or cloud provider). Yes
  \end{enumerate}

  \item If you are using existing assets (e.g., code, data, models) or curating/releasing new assets, check if you include:
  \begin{enumerate}
    \item Citations of the creator If your work uses existing assets. Yes
    \item The license information of the assets, if applicable. Not Applicable
    \item New assets either in the supplemental material or as a URL, if applicable. Not Applicable
    \item Information about consent from data providers/curators. Not Applicable
    \item Discussion of sensible content if applicable, e.g., personally identifiable information or offensive content. Not Applicable
  \end{enumerate}

  \item If you used crowdsourcing or conducted research with human subjects, check if you include:
  \begin{enumerate}
    \item The full text of instructions given to participants and screenshots. Not Applicable
    \item Descriptions of potential participant risks, with links to Institutional Review Board (IRB) approvals if applicable. Not Applicable
    \item The estimated hourly wage paid to participants and the total amount spent on participant compensation. Not Applicable
  \end{enumerate}

\end{enumerate}

\clearpage
\appendix
\thispagestyle{empty}

\onecolumn
\aistatstitle{Deconfounding Scores and Representation Learning for Causal Effect Estimation with Weak Overlap: \\
Supplementary Materials}

\section{PROOFS}
\label{app:proofs}

\subsection{Proof of Lemma \ref{lem:cb}}

For any $\rep$, we have
\begin{align*}
    &\tau_{\rep(X)}  - \tau\\
    &=\tau_{\rep(X)}  - \tau_X \text{ by unconfoundedness}\\
    &= \Et\left[Y\right] - \Et\left[m_0(\rep(X))\right] - \left(\Et\left[Y\right] - \Et\left[m_0(X)\right]\right)\\
    &= \Et\left[\Es[Y|X]\right] - \Et\left[\Es[Y|\rep(X)]\right] \\
    &= \Es\left[\left(m_0(X) - \Es[m_0(X) | \rep(X)]\right) \cdot \left(\densityst{X}(X) - \Es\left[\densityst{X}(X) \middle| \rep(X)\right] \right)\right]\\
    & \ \ \ \ \text{ from Proposition 3.4 of \citet{clivio2024trlfwpidbci} and }m_0(\rep(X)) = \Es[m_0(X) | \rep(X)] \\
    &= \Es\left[\Es\left[\left(m_0(X) - \Es[m_0(X) | \rep(X)]\right) \cdot \left(\densityst{X}(X) - \Es\left[\densityst{X}(X) \middle| \rep(X)\right] \right) \middle| \rep(X)\right] \right]\\
    & \ \ \ \ \text{ from the tower property} \\
    &= \Es\left[\text{Cov}_{\Ps}\left(m_0(X), \densityst{X}(X) \middle| \rep(X)\right)\right].
\end{align*}

\subsection{Proof of Lemma \ref{lem:misoverlap}}

Let $\rep$ be a representation. From Theorem 1 of \citet{hahn1998otrotpsieseoate}, $V_{\text{eff}}^{\rep(X)}$ is equal to
\begin{align*}
\E_{\Pall}\Big[&\frac{e(\rep(X))\text{Var}_{\Pt}(Y|\rep(X))}{\pi_1^2} + \frac{e(\rep(X))^2\text{Var}_{\Ps}(Y|\rep(X))}{\pi_1^2(1-e(\rep(X)))}\\
&+ \frac{\left(m_1(\rep(X)) - m_0(\rep(X)) - \tau_{\rep(X)}\right)^2e(\rep(X))}{\pi_1^2}\Big].\\
\end{align*}
We note that the second term of this sum would be equal to the overlap divergence up to a constant if the conditional variance of $Y$ was constant. Thus, we attempt to upper-bound or lower-bound this conditional variance with constants; this will give the result.

From the law of total variance, 
\begin{align*}
    \text{Var}_{\Ps}(Y|\rep(X)) &= \Es[\text{Var}_{\Ps}(Y|\rep(X), X) | \rep(X)] + \text{Var}_{\Ps}(\Es[Y|\rep(X),X]|\rep(X)) \\
    &=  \Es[\text{Var}_{\Ps}(Y|X) | \rep(X)] + \text{Var}_{\Ps}(m_0(X)|\rep(X)) \\
    &\geq \Es[\text{Var}_{\Ps}(Y|X) | \rep(X)], 
\end{align*}
so if $\text{Var}_{\Ps}(Y|X) \geq \sigma^2$ then $\text{Var}_{\Ps}(Y|\rep(X)) \geq \sigma^2$ and 
    \begin{align*}
        V_{\text{eff}}^{\rep(X)} &\geq \E_{\Pall}\left[\frac{e(\rep(X))^2\text{Var}_{\Ps}(Y|\rep(X))}{\pi_1^2(1-e(\rep(X)))}\right]\\
        &\geq \sigma^2\E_{\Pall}\left[\frac{e(\rep(X))^2}{\pi_1^2(1-e(\rep(X)))}\right]\text{ from the above}\\
        &= \sigma^2\E_{\Pall}\left[\frac{1-e(\rep(X))}{(1-\pi_1)^2}\left(\densityst{\rep(X)}(\rep(X))\right)^2\right] \text{ as }\densityst{\rep(X)}(\rep(X)) = \frac{(1-\pi_1)e(\rep(X))}{\pi_1(1-e(\rep(X)))}\\
        &= \frac{\sigma^2}{1-\pi_1}\E_{\Ps}\left[\left(\densityst{\rep(X)}(\rep(X))\right)^2\right] \text{ as }\density{\rep(X)}{\Pall}{\Ps}(\rep(X)) = \frac{1-e(\rep(X))}{1-\pi_1} \\
        &= \frac{\sigma^2}{1-\pi_1}\misoverlapst(\rep(X)).
    \end{align*}
Further, if $|Y| \leq Y_{\text{max}}$ for a constant $Y_{\text{max}} > 0$ then any conditional variance of $Y$ is bounded by $Y_{\text{max}}^2$ and any conditional expectation of $Y$ is bounded by $Y_{\text{max}}$, and the propensity score is always bounded by 1, so
\begin{align*}
\E_{\Pall}\Big[\frac{e(\rep(X))\text{Var}_{\Pt}(Y|\rep(X))}{\pi_1^2}\Big] &= \Et\Big[\frac{\text{Var}_{\Pt}(Y|\rep(X))}{\pi_1}\Big] \text{ as }\density{\rep(X)}{\Pall}{\Pt}(\rep(X)) = \frac{e(\rep(X))}{\pi_1} \\
&\leq \frac{Y_{\text{max}}^2}{\pi_1}
\end{align*}
and, similarly as above,
\begin{align*}
    \E_{\Pall}\Big[ \frac{e(\rep(X))^2\text{Var}_{\Ps}(Y|\rep(X))}{\pi_1^2(1-e(\rep(X)))}\Big] \leq \frac{Y_{\text{max}}^2}{1-\pi_1}\misoverlapst(\rep(X)).\\
\end{align*}
Finally,
\begin{align*}
    &\E_{\Pall}\Big[
\frac{\left(m_1(\rep(X)) - m_0(\rep(X)) - \tau_{\rep(X)}\right)^2e(\rep(X))}{\pi_1^2}\Big]\\
&= \frac{1}{\pi_1} \Et\Big[\left(m_1(\rep(X)) - m_0(\rep(X)) - \tau_{\rep(X)}\right)^2\Big] \text{ as }\density{\rep(X)}{\Pall}{\Pt}(\rep(X)) = \frac{e(\rep(X))}{\pi_1} \\
&= \frac{1}{\pi_1}\text{Var}_{\Pt}\Big(m_1(\rep(X)) - m_0(\rep(X)) \Big) \\
&\leq \frac{1}{\pi_1} \Et\Big[\left(m_1(\rep(X)) - m_0(\rep(X))\right)^2\Big] \\
&\leq \frac{1}{\pi_1} \Et\Big[2\left(m_1(\rep(X))^2 + m_0(\rep(X))^2\right)\Big] \text{ from }(a-b)^2 \leq 2(a^2+b^2) \ \ \forall  a, b \\
&\leq \frac{4Y_{\text{max}}^2}{\pi_1}. \\
\end{align*}
All of this yields
\begin{align*}
    V_{\text{eff}}^{\rep(X)} \leq \frac{5 Y_{\text{max}}^2}{\pi_1} + \frac{Y_{\text{max}}^2}{1-\pi_1}\misoverlapst(\rep(X)).
\end{align*}

\subsection{Proof of Lemma \ref{lem:bse}}

For any $\rep$, we have
\begin{align*}
    &\misoverlapst(X) - \misoverlapst(\rep(X)) \\
    &= \Es\left[\left(\densityst{X}(X)\right)^2\right] - \Es\left[\left(\densityst{\rep(X)}(\rep(X))\right)^2\right] \\
    &= \Es\left[\left(\densityst{X}(X)\right)^2\right] - \Es\left[\Es\left[\densityst{X}(X)  \middle| \rep(X)\right]^2\right]\\
    & \ \ \ \ \ \ \ \ \ \ \text{ from Proposition 3.4 of \citet{clivio2024trlfwpidbci}}\\
    &= \Es\left[\Es\left[\left(\densityst{X}(X)\right)^2 \middle| \rep(X)\right]\right] - \Es\left[\Es\left[\densityst{X}(X)  \middle| \rep(X)\right]^2\right]\\
    & \ \ \ \ \ \ \ \ \ \ \text{ from the tower property} \\
    &=\Es\left[ \Es\left[\left(\densityst{X}(X)\right)^2 \middle| \rep(X)\right] - \Es\left[\densityst{X}(X)  \middle| \rep(X)\right]^2\right] \\
    &= \Es\left[\text{Var}_{\Ps}\left(\densityst{X}(X) \middle| \rep(X)\right)\right].
\end{align*}
Then,
\begin{align*}
    &\left|\tau_{\rep(X)}  - \tau\right|\\
    &= \left|\Es\left[\left(m_0(X) - \Es[m_0(X) | \rep(X)]\right) \cdot \left(\densityst{X}(X) - \Es\left[\densityst{X}(X) \middle| \rep(X)\right] \right)\right]\right|\\
    & \ \ \ \ \text{ from the proof of Lemma \ref{lem:cb}} \\
    &\leq \sqrt{\Es\left[\left(m_0(X) - \Es[m_0(X) | \rep(X)]\right)^2 \right]}\sqrt{\Es\left[ \left(\densityst{X}(X) - \Es\left[\densityst{X}(X) \middle| \rep(X)\right] \right)^2\right]} \\
    & \ \ \ \ \text{ from the Cauchy-Schwarz inequality} \\
    &=    \sqrt{\Es\left[\Es\left[\left(m_0(X) - \Es[m_0(X) | \rep(X)]\right)^2 \middle| \rep(X)\right]\right]}\\
    & \ \ \ \ \times \sqrt{\Es\left[\Es\left[ \left(\densityst{X}(X) - \Es\left[\densityst{X}(X) \middle| \rep(X)\right] \right)^2 \middle| \rep(X)\right]\right]}  \\
    & \ \ \ \ \text{ from the tower property} \\
    &= \sqrt{\Es\left[\text{Var}_{\Ps}\left(m_0(X) \middle| \rep(X)\right)\right]}\sqrt{\Es\left[\text{Var}_{\Ps}\left(\densityst{X}(X) \middle| \rep(X)\right)\right]} \\
    &= \sqrt{\Es\left[\text{Var}_{\Ps}\left(m_0(X) \middle| \rep(X)\right)\right]}\sqrt{\misoverlapst(X) - \misoverlapst(\rep(X))}  \text{ from the above}.
\end{align*}

\subsection{Proof of Theorem \ref{th:hyperbola}}

\paragraph{Preliminary.} Assume that $R$ is a distribution such that $R_X = \mathcal{N}(0,I_d)$. We show that for any functions $f_1,f_2$ and any $\gamma \in \mathcal{D}_{\alpha,\beta}$, then $\text{Cov}_R\left(f_1(\alpha'X), f_2(\beta'X) | \gamma'X \right) = 0$. This result will be used to prove the Theorem by expressing the confounding bias using such a conditional covariance.

Indeed, as $R_X = \mathcal{N}(0,I_d)$, $\text{Cov}_R\left(\alpha'X, \beta'X | \gamma'X \right) = 0$ implies $\text{Cov}_R\left(f_1(\alpha'X), f_2(\beta'X) | \gamma'X \right) = 0$ regardless of $f_1,f_2$, and $\text{Cov}_R\left(\alpha'X, \beta'X | \gamma'X \right) = \alpha'\beta - \alpha'\gamma \gamma'\beta$ so $\text{Cov}_R\left(\alpha'X, \beta'X | \gamma'X \right)$  being zero is equivalent to

\begin{align*}
\gamma' \left(\frac{\alpha \beta' + \beta \alpha'}{2}\right)\gamma = \alpha'\beta.
\end{align*}

 As $\alpha'\beta \notin \{-1,1\}$ with $||\alpha||_2 = ||\beta||_2 = 1$, $\alpha$ and $\beta$ are not collinear so $\left(\frac{\alpha \beta' + \beta \alpha'}{2}\right)$ is rank 2 with exactly one positive eigenvalue and one negative eigenvalue.  
 Specifically, the eigenvectors $u_1, u_2$ and eigenvalues $\lambda_1, \lambda_2$ are given by
    
 \begin{align}
 u_1 &= \frac{(\alpha + \beta)}{\sqrt{2 + 2\alpha'\beta}}, u_2 = \frac{(\alpha - \beta)}{\sqrt{2 - 2\alpha'\beta}},\\
 \lambda_1 &= \frac{(\alpha'\beta+1)}{2}, \lambda_2 = \frac{(\alpha'\beta-1)}{2}.
 \end{align}

 Then we have a solution characterized by the hyperbola
 \begin{equation}
 \label{eq:hyperbola constraint}
 (\alpha'\beta + 1)\left(\frac{(\alpha + \beta)'\gamma}{\sqrt{2 + 2\alpha'\beta}}\right)^2 - (1 - \alpha'\beta)\left(\frac{(\alpha - \beta)'\gamma}{\sqrt{2 - 2\alpha'\beta}}\right)^2 = 2\alpha'\beta.
 \end{equation}
 Simplifying notation by collapsing scalars into $w_1$ and $w_2$ yields that for any $\gamma \in \mathcal{D}_{\alpha,\beta}$, we have $\text{Cov}_R\left(f_1(\alpha'X), f_2(\beta'X) | \gamma'X \right) = 0$. We now use this to prove the result, separating by Assumption \ref{ass:lineargaussian1} or \ref{ass:lineargaussian2}.

\paragraph{If Assumption \ref{ass:lineargaussian1} Applies.} From Lemma \ref{lem:cb}, a representation $\rep_\gamma(X)$ will be a deconfounding score if
\begin{align*}
    0 = \tau_{\rep_\gamma(X)} - \tau&= \Es\left[\text{Cov}_{\Ps}\left(m_0(X), \densityst{X}(X) \middle| \rep_\gamma(X)\right)\right]\\
    &= \Es\left[\text{Cov}_{\Ps}\left(m(\alpha'X), h(\beta'X) | \gamma'X\right)\right]
\end{align*}
where $\Ps_X = \mathcal{N}(0,I_d)$ so the Preliminary gives the result.

\paragraph{If Assumption \ref{ass:lineargaussian2} Applies.} From Proposition 3.4 of \citet{clivio2024trlfwpidbci}, the confounding bias can be written as
\begin{align*}
    &\tau_{\rep_\gamma(X)} - \tau\\
    &= \Es\left[m_0(X)\left(\densityst{X}(X) - \densityst{\rep_\gamma(X)}(\rep_\gamma(X))\right)\right]\\
    &= \frac{1-\pi_1}{\pi_1}\Es\left[m_0(X)\left(\frac{e(X)}{1-e(X)} - \frac{e(\rep_\gamma(X))}{1-e(\rep_\gamma(X))}\right)\right]\\
    &= \frac{1}{\pi_1}\Es\left[\frac{1-\pi_1}{1-e(X)}m_0(X)\left(e(X) - e(\rep_\gamma(X))\right)\right]\\
    & \ \ \ \ \ \ + \frac{1}{\pi_1}\Es\left[\frac{1-\pi_1}{1-e(X)}m_0(X)\left(e(\rep_\gamma(X)) - \frac{e(\rep_\gamma(X))(1-e(X))}{1-e(\rep_\gamma(X))}\right)\right] \\
    &= \frac{1}{\pi_1}\E_{\Pall}\left[m_0(X)\left(e(X) - e(\rep_\gamma(X))\right)\right]\\
    & \ \ \ \ \ \ + \frac{1}{\pi_1}\E_{\Pall}\left[m_0(X)\left(e(\rep_\gamma(X)) - \frac{e(\rep_\gamma(X))(1-e(X))}{1-e(\rep_\gamma(X))}\right)\right] \text{ as }\density{X}{\Ps}{\Pall}(X) = \frac{1-\pi_1}{1-e(X)} \\
    &= \frac{1}{\pi_1}\E_{\Pall}\left[m_0(X)\left(e(X) - e(\rep_\gamma(X))\right)\right]\\
    & \ \ \ \ \ \ + \frac{1}{\pi_1}\E_{\Pall}\left[\frac{e(\rep_\gamma(X))}{1-e(\rep_\gamma(X))}m_0(X)\Big(1-e(\rep_\gamma(X)) -(1-e(X))\Big)\right] \\
    &= \frac{1}{\pi_1}\E_{\Pall}\left[m_0(X)\left(e(X) - e(\rep_\gamma(X))\right)\right]\\
    & \ \ \ \ \ \ + \frac{1}{\pi_1}\E_{\Pall}\left[\frac{e(\rep_\gamma(X))}{1-e(\rep_\gamma(X))}m_0(X)\Big(e(X) - e(\rep_\gamma(X))\Big)\right].
\end{align*}
Further, from the tower property, for any functions $f_1, f_2$ of $X$ and $f_3$ of $\rep_\gamma(X)$,
\begin{align*}
    &\mathbb{E}_{\Pall}\left[f_3(\rep_\gamma(X)) \cdot f_1(X) \cdot \left(f_2(X) - \mathbb{E}_{\Pall}\left[f_2(X) | \rep_\gamma(X)\right]\right)\right]\\
    &= \mathbb{E}_{\Pall}\left[f_3(\rep_\gamma(X)) \cdot \left(f_1(X) - \mathbb{E}_{\Pall}\left[f_1(X) | \rep_\gamma(X)\right]\right) \cdot \left(f_2(X) - \mathbb{E}_{\Pall}\left[f_2(X) | \rep_\gamma(X)\right]\right)\right] \\ 
    &= \mathbb{E}_{\Pall}\left[f_3(\rep_\gamma(X)) \cdot\mathbb{E}_{\Pall}\left[\left(f_1(X) - \mathbb{E}_{\Pall}\left[f_1(X) | \rep_\gamma(X)\right]\right) \cdot \left(f_2(X) - \mathbb{E}_{\Pall}\left[f_2(X) | \rep_\gamma(X)\right]\right)\right] | \rep_\gamma(X)\right] \\
    &= \mathbb{E}_{\Pall}\left[f_3(\rep_\gamma(X)) \cdot\text{Cov}_\Pall\left(f_1(X), f_2(X) | \rep_\gamma(X)\right)\right]
\end{align*}
and, as $e(\rep_\gamma(X)) = \mathbb{E}_\Pall\left[e(X) | \rep_\gamma(X)\right]$, we obtain
\begin{align*}
    &\tau_{\rep_\gamma(X)} - \tau \\
    &= \frac{1}{\pi_1}\E_{\Pall}\left[\text{Cov}_{\Pall}\left(m_0(X), e(X) | \rep_\gamma(X)\right)\right] + \frac{1}{\pi_1}\E_{\Pall}\left[\frac{e(\rep_\gamma(X))}{1-e(\rep_\gamma(X))}\text{Cov}_{\Pall}\left(m_0(X), e(X) | \rep_\gamma(X)\right)\right] \\
    &= \frac{1}{\pi_1}\E_{\Pall}\left[\text{Cov}_{\Pall}\left(m(\alpha'X), h(\beta'X) | \gamma'X\right)\right]\\
    & \ \ \ \ \ \ + \frac{1}{\pi_1}\E_{\Pall}\left[\frac{e(\gamma'X)}{1-e(\gamma'X)}\text{Cov}_{\Pall}\left(m(\alpha'X), h(\beta'X) | \gamma'X\right)\right],
\end{align*}
so the Preliminary applied to $\Pall_X = \mathcal{N}(0,I_d)$ gives the result.

\subsection{Proof of Theorem \ref{th:misoverlap_prognostic_score}}

\paragraph{Sketch of the Proof.} First, we compute the global minimizers of $\gamma \mapsto |\beta'\gamma|$; then we show that the overlap divergence is a non-decreasing or increasing function of $|\beta'\gamma|$, showing that the former global minimizers of $|\beta'\gamma|$ are also (the unique) global minimizers of the overlap divergence.

\subsubsection{Optimal $\gamma$'s for $|\beta'\gamma|$}

We derive them by separating the study depending on the orientation of the hyperbola. Let $\gamma \in \mathcal{D}_{\alpha, \beta}$ with associated $w_1, w_2, n$.

\paragraph{Assume that $\alpha'\beta \geq 0$.} First, note that reparameterizing $w_1, w_2$ as
\begin{align*}
    w_1 &= \epsilon_a \sqrt{\frac{2\alpha'\beta + (1-\alpha'\beta)w^2}{1 + \alpha'\beta}}, \\
    w_2 &= \epsilon_b w,
\end{align*}
with $\epsilon_a, \epsilon_b \in \{-1,1\}, w \in \left[0, \sqrt{\frac{1-\alpha'\beta}{2}}\right]$, we have
\begin{align*}
    \gamma &= \frac{\epsilon_a\sqrt{2\alpha'\beta + (1 - \alpha'\beta)w^2}\cdot(\alpha + \beta)}{\sqrt{2}(1 + \alpha'\beta)} +   \frac{\epsilon_bw\cdot(\alpha - \beta)}{\sqrt{2(1-\alpha'\beta)}} + \sqrt{\frac{1 - \alpha'\beta - 2w^2}{1 + \alpha'\beta}} \cdot n,\\
    \beta'\gamma &= \frac{1}{\sqrt{2}}\left(\epsilon_a\sqrt{2\alpha'\beta + (1-\alpha'\beta)w^2} - \epsilon_bw\sqrt{1-\alpha'\beta}\right),
\end{align*}
where the term with $n$ has been removed as $\beta'n = 0$ by definition of $n$.

\paragraph{Assume that $\alpha'\beta \leq 0$.} First, note that reparameterizing $w_1, w_2$ as
\begin{align*}
    w_2 &= \epsilon_a \sqrt{\frac{-2\alpha'\beta + (1+\alpha'\beta)w^2}{1 - \alpha'\beta}}, \\
    w_1 &= \epsilon_b w,
\end{align*}
with $\epsilon_a, \epsilon_b \in \{-1,1\}, w \in \left[0, \sqrt{\frac{1+\alpha'\beta}{2}}\right]$, we have
\begin{align*}
    \gamma &= \frac{\epsilon_a\sqrt{-2\alpha'\beta + (1 + \alpha'\beta)w^2}\cdot(\alpha - \beta)}{\sqrt{2}(1 - \alpha'\beta)} +   \frac{\epsilon_bw\cdot(\alpha + \beta)}{\sqrt{2(1+\alpha'\beta)}} + \sqrt{\frac{1 + \alpha'\beta - 2w^2}{1 - \alpha'\beta}} \cdot n,\\
    \beta'\gamma &= \frac{1}{\sqrt{2}}\left(-\epsilon_a\sqrt{-2\alpha'\beta + (1+\alpha'\beta)w^2} + \epsilon_bw\sqrt{1+\alpha'\beta}\right),
\end{align*}
where the term with $n$ has been removed as $\beta'n = 0$ by definition of $n$.

\paragraph{Factorization.} Thus, noting $\text{sign}(x) = 2\cdot1_{\{x \geq 0\}} - 1$ we always have
\begin{align*}
    \gamma &= \frac{\epsilon_a\sqrt{2|\alpha'\beta| + (1 - |\alpha'\beta|)w^2}\cdot(\alpha + \text{sign}(\alpha'\beta)\beta)}{\sqrt{2}(1 + |\alpha'\beta|)}\\
    & \ \  \ \ \ \ \ \ \ \ +   \frac{\epsilon_bw\cdot(\alpha - \text{sign}(\alpha'\beta)\beta)}{\sqrt{2(1-|\alpha'\beta|)}} + \sqrt{\frac{1 - |\alpha'\beta| - 2w^2}{1 + |\alpha'\beta|}} \cdot n,\\
    \beta'\gamma &= \frac{\text{sign}(\alpha'\beta)}{\sqrt{2}}\left(\epsilon_a\sqrt{2|\alpha'\beta| + (1-|\alpha'\beta|)w^2} - \epsilon_bw\sqrt{1-|\alpha'\beta|}\right),
\end{align*}
with $\epsilon_a, \epsilon_b \in \{-1,1\}, w \in \left[0, \sqrt{\frac{1-|\alpha'\beta|}{2}}\right]$.
Then,
\begin{align*}
    (\beta'\gamma)^2
    &= \frac{1}{2}\Big(2|\alpha'\beta| + (1-|\alpha'\beta|)w^2 + w^2(1-|\alpha'\beta|)\\
    & \  \ \ \ \ \ \ \ \  \ \ \ \  - 2\epsilon_a\epsilon_bw\sqrt{1-|\alpha'\beta|}\sqrt{2|\alpha'\beta| + (1-|\alpha'\beta|)w^2}\Big) \\
    &= |\alpha'\beta| + z - \epsilon_a\epsilon_b\sqrt{2|\alpha'\beta| z + z^2},
\end{align*}
where $z := (1 - |\alpha'\beta|)w^2 \in \left[0, \frac{(1 - |\alpha'\beta|)^2}{2}\right]$. From there, note that if $\epsilon_a\epsilon_b = -1$ then $(\beta'\gamma)^2$ is an increasing function of $z$: thus a unique global minimizer in this case is $z = 0$, for which $(\beta'\gamma)^2 = |\alpha'\beta|$. Now assume that $\epsilon_a\epsilon_b = 1$. Then,
\begin{align*}
    \frac{\partial (\beta'\gamma)^2}{\partial z} = 1 - \frac{|\alpha'\beta| + z}{\sqrt{2|\alpha'\beta| z + z^2}},
\end{align*}
whose sign is given by $2|\alpha'\beta| z + z^2 - (|\alpha'\beta| + z)^2 = -|\alpha'\beta|^2$.

Thus, for $\epsilon_a\epsilon_b > 0$, $(\beta'\gamma)^2$ is a decreasing function of $z$ if $\alpha'\beta \neq 0$ and $(\beta'\gamma)^2$ is constant if $\alpha'\beta = 0$. Then one global minimum, and the only one if $\alpha'\beta \neq 0$, is achieved for the maximal value of $z$, that is $\frac{(1-|\alpha'\beta|)^2}{2}$. For this value of $z$,
\begin{align*}
    (\beta'\gamma)^2 &= |\alpha'\beta| + \frac{(1-|\alpha'\beta|)^2}{2} - \frac{1-|\alpha'\beta|}{\sqrt{2}}\sqrt{2|\alpha'\beta| + \frac{(1-|\alpha'\beta|)^2}{2}} \\
    &= \frac{1 + |\alpha'\beta|^2}{2} - \frac{(1 - |\alpha'\beta|)(1 + |\alpha'\beta|)}{2} \\
    &= |\alpha'\beta|^2
\end{align*}
which is lower than $|\alpha'\beta|$. Thus,
\begin{itemize}
    \item If $\alpha'\beta \neq 0$, $(\beta'\gamma)^2$ is minimized for $\epsilon_a = \epsilon_b =: \epsilon \in \{-1,1\}$ and $w = \sqrt{\frac{1 - |\alpha'\beta|}{2}}$, yielding
\begin{align*}
    \gamma &=  \epsilon\frac{\sqrt{2|\alpha'\beta| + \frac{(1 - |\alpha'\beta|)^2}{2}}}{\sqrt{2}(1 + |\alpha'\beta|)}(\alpha + \text{sign}(\alpha'\beta) \beta) + \epsilon \sqrt{\frac{1 - |\alpha'\beta|}{2}} \frac{\alpha - \text{sign}(\alpha'\beta)\beta}{\sqrt{2(1-|\alpha'\beta|)}}\\
    & \ \ \  \ \ \ \ \ \ \ \ \ \text{ where the term in $n$ is zero}\\
    &= \frac{\epsilon}{2}(\alpha + \text{sign}(\alpha'\beta)\beta) + \frac{\epsilon}{2}(\alpha-\text{sign}(\alpha'\beta)\beta) \\
    &= \epsilon\alpha.
\end{align*}
\item If $\alpha'\beta = 0$, $(\beta'\gamma)^2$ is minimized for $\epsilon_a = \epsilon_b =: \epsilon \in \{-1,1\}$ and any $w$, yielding
\begin{align*}
    \gamma &= \frac{\epsilon}{\sqrt{2}}w(\alpha + \text{sign}(\alpha'\beta)\beta) + \frac{\epsilon}{\sqrt{2}}w(\alpha - \text{sign}(\alpha'\beta)\beta) + \sqrt{1-2w^2}n \\
    &= \sqrt{2}\epsilon w \alpha + \sqrt{1-2w^2}n.
\end{align*}
As $\epsilon \in \{-1,1\}$, and $w \in [0,\frac{1}{\sqrt{2}}]$, we have that $\sqrt{2}\epsilon w \alpha$ spans the entire segment $[-\alpha,\alpha]$.
\end{itemize}

\paragraph{Conclusion.} As $\epsilon_a = 1$ corresponds to the portion of $\mathcal{D}_{\alpha,\beta}$ with endpoints $\alpha$ and $\text{sign}(\alpha'\beta)\beta$, and $\epsilon_a = -1$ to that with endpoints $-\alpha$ and $-\text{sign}(\alpha'\beta)\beta$, we have that 

\begin{itemize}
\item If $\alpha'\beta \neq 0$, $|\beta'\gamma|$ is decreasing when moving from $\text{sign}(\alpha'\beta)\beta$ to $\alpha$ on the portion of $\mathcal{D}_{\alpha,\beta}$ with endpoints $\alpha$ and $\text{sign}(\alpha'\beta)\beta$ or when moving from $-\text{sign}(\alpha'\beta)\beta$ to $-\alpha$ on the portion of $\mathcal{D}_{\alpha,\beta}$ with endpoints $-\alpha$ and $-\text{sign}(\alpha'\beta)\beta$; notably, $\gamma = \alpha$ and $\gamma = -\alpha$ are the only global minimizers of $|\beta'\gamma|$ on $\mathcal{D}_{\alpha,\beta}$.
\item If $\alpha'\beta = 0$,  the $\gamma$'s whose projection on the span of $\alpha$ and $\beta$ belongs to the segment $[-\alpha,\alpha]$ are the only global optimizers of $|\beta'\gamma|$ on $\mathcal{D}_{\alpha,\beta}$.
\end{itemize}

Thus, the proof of Theorem \ref{th:misoverlap_prognostic_score} consists in proving that $\misoverlapst(\rep_\gamma(X))$ is a non-decreasing or increasing function of $|\beta'\gamma|$.

\subsubsection{Proof of 1.(a), 1.(b), 2.(a)(i) and 2.(b)}

This part of the proof is done by expressing the overlap divergence in terms of a specific function of $|\beta'\gamma|$ and using the dominated convergence theorem to differentiate this function; the derivative is then shown to be non-negative (or positive).

\paragraph{Expression of the Overlap Divergence in 1.(a) and 2.(a)(i).} For simplicity, and without loss of generality, assume that $\mathbb{E}_{Z \sim \mathcal{N}(0,1)}[h(Z)] = 1$. Also, remember that $\beta'X | \gamma'X \sim_{\Ps} \mathcal{N}((\beta'\gamma)\gamma'X, 1 - (\beta'\gamma)^2)$, so
\begin{align*}
    \Es\left[\densityst{X}(X) \middle| \gamma'X \right] &= \E_{Z \sim \mathcal{N}((\beta'\gamma)\gamma'X, 1 - (\beta'\gamma)^2)}\left[h(Z)\right]\\
    &= \E_{Z' \sim \mathcal{N}(0, 1)}\left[h((\beta'\gamma)\gamma'X + \sqrt{1 - (\beta'\gamma)^2}Z')\right]
\end{align*}
and, as $\gamma'X \sim_{\Ps} \mathcal{N}(0, 1)$,
\begin{align*}
    \misoverlapst(\rep_\gamma(X)) &= \Es\left[\left(\densityst{\rep_\gamma(X)}(\rep_\gamma(X))\right)^2\right] \\
    &= \Es\left[\Es\left[\densityst{X}(X) \middle| \rep_\gamma(X) \right]^2\right] \text{ from Proposition 3.4 of \citet{clivio2024trlfwpidbci}}\\
    &= \Es\left[\Es\left[\densityst{X}(X) \middle| \gamma'X \right]^2\right] \\
    &= \E_{Z \sim \mathcal{N}(0,1)}\left[\Es\left[\densityst{X}(X) \middle| \gamma'X = Z \right]^2\right] \\
    &= \E_{Z \sim \mathcal{N}(0,1)}\left[\E_{Z' \sim \mathcal{N}(0, 1)}\left[h((\beta'\gamma)Z + \sqrt{1 - (\beta'\gamma)^2}Z')\right]^2\right] \\
    &= g(\beta'\gamma)
\end{align*}
where $g(u) := \E_{Z \sim \mathcal{N}(0,1)}\left[\E_{Z' \sim \mathcal{N}(0, 1)}\left[h(uZ + \sqrt{1 - u^2}Z')\right]^2\right]$ for $u \in [-1,1]$.

\paragraph{Expression of the Overlap Divergence in 1.(b) and 2.(b).} Again, $\beta'X | \gamma'X \sim_{\Pall} \mathcal{N}((\beta'\gamma)\gamma'X, 1 - (\beta'\gamma)^2)$, so
\begin{align*}
    e(\rep_\gamma(X))
    &= \E_{\Pall}\left[e(X) | \rep_\gamma(X)\right] \\
    &= \E_{\Pall}\left[h(\beta'X) \middle| \gamma'X \right] \\
    &= \E_{Z \sim \mathcal{N}((\beta'\gamma)\gamma'X, 1 - (\beta'\gamma)^2)}\left[h(Z)\right]\\
    &= \E_{Z' \sim \mathcal{N}(0, 1)}\left[h((\beta'\gamma)\gamma'X + \sqrt{1 - (\beta'\gamma)^2}Z')\right]
\end{align*}
and, as $\rep_\gamma(X) = \gamma'X \sim_{\Pall} \mathcal{N}(0, 1)$,
\begin{align*}
    \misoverlapst(\rep_\gamma(X)) &= \Es\left[\left(\densityst{\rep_\gamma(X)}(\rep_\gamma(X))\right)^2\right] \\
    &= \Es\left[\frac{(1-\pi_1)^2}{\pi_1^2}\frac{e(\rep_\gamma(X))^2}{(1-e(\rep_\gamma(X)))^2}\right] \\
    &= \EP{\Pall}\left[\frac{1-\pi_1}{\pi_1^2}\frac{e(\rep_\gamma(X))^2}{1-e(\rep_\gamma(X))}\right] \\
    &= g(\beta'\gamma)
\end{align*}
where $g(u) := \E_{Z \sim \mathcal{N}(0,1)}\left[\frac{1-\pi_1}{\pi_1^2}\frac{\E_{Z' \sim \mathcal{N}(0, 1)}\left[h(uZ + \sqrt{1 - u^2}Z')\right]^2}{1 - \E_{Z' \sim \mathcal{N}(0, 1)}\left[h(uZ + \sqrt{1 - u^2}Z')\right]}\right]$ for $u \in [-1,1]$.

\paragraph{Factorization of 1.(a), 1.(b), 2.(a)(i) and 2.(b).} Thus, in these four settings,
\begin{align*}
    \misoverlapst(\rep_\gamma(X)) = g(\beta'\gamma) \text{ with } g(u) :=  \E_{Z \sim \mathcal{N}(0,1)}\left[f\left(\E_{Z' \sim \mathcal{N}(0, 1)}\left[h(uZ + \sqrt{1 - u^2}Z')\right]\right)\right]
\end{align*}
where $f(t)  = f_a(t) := t^2$ for 1.(a) and 2.(a)(i), and $f(t) = f_b(t) := \frac{1-\pi_1}{\pi_1^2}\frac{t^2}{1-t}$ for 1.(b) and 2.(b).

First, note that $g$ is symmetric as for any $u \in [-1,1]$,
\begin{align*}
    g(-u) &= \E_{Z \sim \mathcal{N}(0,1)}\left[f\left(\E_{Z' \sim \mathcal{N}(0, 1)}\left[h((-u)Z + \sqrt{1 - (-u)^2}Z')\right]\right)\right]\\
    &= \E_{Z \sim \mathcal{N}(0,1)}\left[f\left(\E_{Z' \sim \mathcal{N}(0, 1)}\left[h(u(-Z) + \sqrt{1 - u^2}Z')\right]\right)\right]\\
    &= \E_{Z'' \sim \mathcal{N}(0,1)}\left[f\left(\E_{Z' \sim \mathcal{N}(0, 1)}\left[h(uZ'' + \sqrt{1 - u^2}Z')\right]\right)\right]\\
    & \ \ \ \ \ \ \ \text{ as if $Z \sim \mathcal{N}(0,1)$ then $-Z \sim \mathcal{N}(0,1)$}  \\
    &= g(u).
\end{align*}
Then, for any $u \in [-1,1], g(u) = g(|u|)$. Thus, we need to show that the restriction of $g$ to $[0,1]$ is non-decreasing, or increasing depending on whether we place ourselves in setups 1.(a) and 1.(b) or setups 2.(a)(i) and 2.(b). We show that it is non-decreasing (resp. increasing) on every interval $[0,u_0]$ where $u_0 \in [0,1)$ ; then it will be so on $[0,1)$, and the proof of 1.(a) and 1.(b) (resp. 2.(a)(i) and 2.(b)) is then complete if we show that $\forall u \in [0,1], \ \ g(u) \leq g(1)$. Indeed, let $u \in [0,1]$, 
\begin{align*}
    g(u) &= \E_{Z \sim \mathcal{N}(0,1)}\left[f\left(\E_{Z' \sim \mathcal{N}(0, 1)}\left[h(uZ + \sqrt{1 - u^2}Z')\right]\right)\right] \\
    &\leq \E_{Z \sim \mathcal{N}(0,1)}\left[\E_{Z' \sim \mathcal{N}(0, 1)}\left[f(h(uZ + \sqrt{1 - u^2}Z'))\right]\right]\\
    & \ \  \ \ \ \ \text{ from Jensen's inequality, as $f$ is convex} \\
    &= \E_{Z, Z' \overset{\text{indep}}{\sim} \mathcal{N}(0,1)}\left[f(h(uZ + \sqrt{1 - u^2}Z'))\right] \text{ where }uZ + \sqrt{1 - u^2}Z' \sim \mathcal{N}(0,1) \\
    &= \E_{Z \sim \mathcal{N}(0,1)}\left[f(h(Z))\right] \\
    &= g(1)
\end{align*}
which then completes the proof. We now fix $u_0 \in [0,1)$. In the following, unless specified otherwise, $Z,Z' \overset{\text{indep}}{\sim} \mathcal{N}(0,1)$.\\

\paragraph{Preliminary Bounds.} First we prove that for any $\mu$, $\sigma > 0$ and $\bar{\lambda} < \frac{1}{2\sigma^2}$,
\begin{align*}
    \E_Z\left[|Z|e^{\bar{\lambda}(\mu+\sigma Z)^2}\right] \leq e^{\frac{\bar{\lambda} \mu^2}{1-2\bar{\lambda} \sigma^2}}\left(\frac{1}{\sqrt{1-2\bar{\lambda} \sigma^2}} + \frac{1}{\sqrt{1-2\bar{\lambda} \sigma^2}^3} + \frac{4\bar{\lambda}^2\sigma^2\mu^2}{\sqrt{1-2\bar{\lambda} \sigma^2}^5}\right).
\end{align*}
Indeed, as $\forall z, |z| \leq 1 + z^2$, we have
\begin{align*}
    \E_Z\left[|Z|e^{\bar{\lambda}(\mu+\sigma Z)^2}\right] &\leq \E_Z\left[\left(1 + Z^2\right)e^{\bar{\lambda}(\mu+\sigma Z)^2}\right]\\
    &= \E_Z\left[e^{\bar{\lambda}(\mu+\sigma Z)^2}\right] + \E_Z\left[Z^2e^{\bar{\lambda}(\mu+\sigma Z)^2}\right]. 
\end{align*}
From the MGF of an uncentered $\chi^2$ distribution,
\begin{align*}
    \E_Z\left[e^{\bar{\lambda}(\mu+\sigma Z)^2}\right] = \frac{1}{\sqrt{1-2\bar{\lambda} \sigma^2}}e^{\frac{\bar{\lambda} \mu^2}{1-2\bar{\lambda} \sigma^2}}.
\end{align*}
Then,
\begin{align*}
    \E_Z\left[Z^2e^{\bar{\lambda}(\mu+\sigma Z)^2}\right] = \E_{X \sim \mathcal{N}(\mu,\sigma^2)}\left[\left(\frac{X-\mu}{\sigma}\right)^2e^{\bar{\lambda}X^2}\right]  = \frac{1}{\sigma^2}\Big(&\E_{X \sim \mathcal{N}(\mu,\sigma^2)}\left[X^2e^{\bar{\lambda} X^2}\right] - 2\mu\E_{X \sim \mathcal{N}(\mu,\sigma^2)}\left[Xe^{\bar{\lambda} X^2}\right] \\
    &+ \mu^2\E_{X \sim \mathcal{N}(\mu,\sigma^2)}\left[e^{\bar{\lambda} X^2}\right]\Big)
\end{align*}
where, again,
\begin{align*}
    \E_{X \sim \mathcal{N}(\mu,\sigma^2)}\left[e^{\bar{\lambda} X^2}\right] = \frac{1}{\sqrt{1-2\bar{\lambda} \sigma^2}}e^{\frac{\bar{\lambda} \mu^2}{1-2\bar{\lambda} \sigma^2}}
\end{align*}
and, when $\bar{\lambda} \neq 0$ (the following result trivially holds for $\bar{\lambda} = 0$),
\begin{align*}
    \E_{X \sim \mathcal{N}(\mu,\sigma^2)}\left[Xe^{\bar{\lambda} X^2}\right] &= \E_Z\left[(\mu+\sigma Z)e^{\bar{\lambda}(\mu+\sigma Z)^2}\right] \\
    &=  \frac{1}{2\bar{\lambda}}\E_Z\left[\frac{\partial}{\partial \mu}\left(e^{\bar{\lambda}(\mu+\sigma Z)^2}\right)\right] \\
    &= \frac{1}{2\bar{\lambda}}\frac{\partial}{\partial \mu}\left(\E_Z\left[e^{\bar{\lambda}(\mu+\sigma Z)^2}\right]\right) \\
    &= \frac{1}{2\bar{\lambda}}\frac{\partial}{\partial \mu}\left(\frac{1}{\sqrt{1-2\bar{\lambda} \sigma^2}}e^{\frac{\bar{\lambda} \mu^2}{1-2\bar{\lambda} \sigma^2}}\right) \\
    &= \frac{\mu}{\sqrt{1-2\bar{\lambda} \sigma^2}^3}e^{\frac{\bar{\lambda} \mu^2}{1-2\bar{\lambda} \sigma^2}}
\end{align*}
where differentiation and expectation can be exchanged thanks to the dominated convergence theorem, restricting $\mu$ to $[-M,M]$ for $M > 0$ so $\frac{\partial}{\partial \mu}\left( e^{\bar{\lambda}(\mu+\sigma Z)^2} \right)$ is bounded by $2\bar{\lambda}(M + \sigma |Z|)e^{\bar{\lambda}(M + \sigma |Z|)^2}$ which is integrable, and finally,
\begin{align*}
    \E_{X \sim \mathcal{N}(\mu,\sigma^2)}\left[X^2e^{\bar{\lambda} X^2}\right] &= \E_{X \sim \mathcal{N}(\mu,\sigma^2)}\left[\frac{\partial}{\partial \bar{\lambda}}\left(e^{\bar{\lambda} X^2}\right)\right] \\
    &= \frac{\partial}{\partial \bar{\lambda}}\left(\E_{X \sim \mathcal{N}(\mu,\sigma^2)}\left[e^{\bar{\lambda} X^2}\right]\right) \\
    &= \frac{\partial}{\partial \bar{\lambda}}\left(\frac{1}{\sqrt{1-2\bar{\lambda} \sigma^2}}e^{\frac{\bar{\lambda} \mu^2}{1-2\bar{\lambda} \sigma^2}}\right) \\
    &= \left(\frac{\sigma^2}{\sqrt{1-2\bar{\lambda} \sigma^2}^3} + \frac{\mu^2}{\sqrt{1-2\bar{\lambda} \sigma^2}^5}\right) \cdot e^{\frac{\bar{\lambda} \mu^2}{1-2\bar{\lambda} \sigma^2}}
\end{align*}
where differentiation and expectation can be exchanged thanks to the dominated convergence theorem, restricting $\bar{\lambda}$ to $(-\infty,\bar{\lambda}^\star]$ for $\bar{\lambda}^\star < \frac{1}{2\sigma^2}$ so $\frac{\partial}{\partial \bar{\lambda}}\left(e^{\bar{\lambda} X^2}\right)$ is bounded by $X^2e^{\bar{\lambda}^\star X^2}$ which is integrable. In the end,
\begin{align*}
    \E_Z\left[Z^2e^{\bar{\lambda}(\mu+\sigma Z)^2}\right] &= \frac{1}{\sigma^2}\Big(\E_{X \sim \mathcal{N}(\mu,\sigma^2)}\left[X^2e^{\bar{\lambda} X^2}\right] - 2\mu\E_{X \sim \mathcal{N}(\mu,\sigma^2)}\left[Xe^{\bar{\lambda} X^2}\right]\\
    & \ \ \ \ \ \ \ \ \ \ \ \ \ \ \ \ \ \ + \mu^2\E_{X \sim \mathcal{N}(\mu,\sigma^2)}\left[e^{\bar{\lambda} X^2}\right]\Big) \\
    &= \Big(\frac{\sigma^2}{\sqrt{1-2\bar{\lambda} \sigma^2}^3} + \frac{\mu^2}{\sqrt{1-2\bar{\lambda} \sigma^2}^5} - \frac{2\mu^2}{\sqrt{1-2\bar{\lambda} \sigma^2}^3}\\
    & \ \ \ \ \ \  \ + \frac{\mu^2}{\sqrt{1-2\bar{\lambda} \sigma^2}}\Big) \cdot \frac{e^{\frac{\bar{\lambda} \mu^2}{1-2\bar{\lambda} \sigma^2}}}{\sigma^2} \\
    &= \left(\frac{\sigma^2 - 2\mu^2}{\sqrt{1-2\bar{\lambda} \sigma^2}^3} + \frac{\mu^2}{\sqrt{1-2\bar{\lambda} \sigma^2}^5} + \frac{\mu^2}{\sqrt{1-2\bar{\lambda} \sigma^2}}\right) \cdot \frac{e^{\frac{\bar{\lambda} \mu^2}{1-2\bar{\lambda} \sigma^2}}}{\sigma^2} \\
    &= \left(\frac{\sigma^2 - \mu^2 - 2\bar{\lambda}\sigma^2\mu^2}{\sqrt{1-2\bar{\lambda} \sigma^2}^3} + \frac{\mu^2}{\sqrt{1-2\bar{\lambda} \sigma^2}^5}\right) \cdot \frac{e^{\frac{\bar{\lambda} \mu^2}{1-2\bar{\lambda} \sigma^2}}}{\sigma^2} \\
        &= \left(\frac{\left(\sigma^2 - \mu^2 - 2\bar{\lambda}\sigma^2\mu^2\right)\left(1-2\bar{\lambda} \sigma^2\right)}{\sqrt{1-2\bar{\lambda} \sigma^2}^5} + \frac{\mu^2}{\sqrt{1-2\bar{\lambda} \sigma^2}^5}\right) \cdot \frac{e^{\frac{\bar{\lambda} \mu^2}{1-2\bar{\lambda} \sigma^2}}}{\sigma^2} \\
    &= \left(\frac{\sigma^2\left(1 - 2\bar{\lambda}\sigma^2 + 4\bar{\lambda}^2\mu^2\sigma^2\right)}{\sqrt{1-2\bar{\lambda} \sigma^2}^5}\right) \cdot \frac{e^{\frac{\bar{\lambda} \mu^2}{1-2\bar{\lambda} \sigma^2}}}{\sigma^2} \\
    &= \left( \frac{1}{\sqrt{1-2\bar{\lambda} \sigma^2}^3} + \frac{4\bar{\lambda}^2\mu^2\sigma^2}{\sqrt{1-2\bar{\lambda} \sigma^2}^5}\right) \cdot e^{\frac{\bar{\lambda} \mu^2}{1-2\bar{\lambda} \sigma^2}}
\end{align*}
so the final bound is
\begin{align*}
    \E_Z\left[|Z|e^{\bar{\lambda}(\mu+\sigma Z)^2}\right] &\leq \E_Z\left[e^{\bar{\lambda}(\mu+\sigma Z)^2}\right] + \E_Z\left[Z^2e^{\bar{\lambda}(\mu+\sigma Z)^2}\right]\\
    &= \left(\frac{1}{\sqrt{1-2\bar{\lambda} \sigma^2}} + \frac{1}{\sqrt{1-2\bar{\lambda} \sigma^2}^3} + \frac{4\bar{\lambda}^2\mu^2\sigma^2}{\sqrt{1-2\bar{\lambda} \sigma^2}^5}\right) \cdot e^{\frac{\bar{\lambda} \mu^2}{1-2\bar{\lambda} \sigma^2}}.
\end{align*}
Further, for $u \in [0,u_0] \subset [0,1)$ and $z \in \mathbb{R}$, assume $\mu = uz$, $\sigma^2 = 1 - u^2$, $0 \leq \lambda < \frac{1}{4}$, $0 \leq \lambda' < \frac{1 - 4\lambda}{6}$. Then,
\begin{align*}
    e^{\frac{\lambda \mu^2}{1-2\lambda \sigma^2}} = e^{\frac{\lambda u^2z^2}{1-2\lambda(1-u^2)}} \leq e^{\lambda z^2}
\end{align*}
which follows from the function $u \rightarrow \frac{\lambda u^2z^2}{1-2\lambda(1-u^2)}$ which is non-decreasing when $0 \leq \lambda \leq 1/2$, thus bounded above by $\lambda z^2$. Further, $\frac{1}{\sqrt{1-2\lambda\sigma^2}} \leq \sqrt{2}$. In the end,
\begin{align*}
    &\E_{Z}\left[|Z|e^{\lambda(\mu+\sigma Z)^2}\right] =  \E_{Z}\left[|Z|e^{\lambda(uz+\sqrt{1-u^2}Z)^2}\right] \leq \sqrt{2}(3 + 16\lambda^2z^2) \cdot e^{\lambda z^2},  \\
    &\E_{Z}\left[e^{\lambda(\mu+\sigma Z)^2}\right] = \E_{Z}\left[e^{\lambda(uz+\sqrt{1-u^2}Z)^2}\right] \leq \sqrt{2} \cdot e^{\lambda z^2}.
\end{align*}
Thus, as $\mathcal{H}'^2_{C,\lambda,\lambda'} \subset \mathcal{H}^2_{C,\lambda}$, for any $k = 0,1,2$,
\begin{align*}
    &\E_{Z}\left[\left|Zh^{(k)}(uz+\sqrt{1-u^2}Z)\right|\right] \leq \E_{Z}\left[C|Z|e^{\lambda(uz+\sqrt{1-u^2}Z)^2}\right] \leq \sqrt{2}(3 + 16\lambda^2z^2)C \cdot e^{\lambda z^2},  \\
    &\E_{Z}\left[\left|h^{(k)}(uz+\sqrt{1-u^2}Z)\right|\right] \leq \E_{Z}\left[Ce^{\lambda(uz+\sqrt{1-u^2}Z)^2}\right] \leq \sqrt{2}C \cdot e^{\lambda z^2}.
\end{align*}
Further, if $h \in \mathcal{H}'^2_{C,\lambda,\lambda'}$, 
\begin{align*}
    \frac{1}{1-\E_{Z}\left[h(uz+\sqrt{1-u^2}Z)\right]} &= \frac{1}{\E_{Z}\left[(1 - h)(uz+\sqrt{1-u^2}Z)\right]}\\
    &\leq \frac{1}{\E_{Z}\left[\frac{1}{C}e^{-\lambda'(uz+\sqrt{1-u^2}Z)^2}\right]} \\
    &= C\sqrt{1+2\lambda'(1-u^2)}e^{\lambda'\frac{u^2z^2}{1+2\lambda'(1-u^2)}} \\
    &\leq \frac{2}{\sqrt{3}}Ce^{\lambda'z^2}
\end{align*}
as $u \mapsto \lambda'\frac{u^2z^2}{1+2\lambda'(1-u^2)}$ is non-decreasing on $[0,1)$. As $f_b'(t) = \frac{1-\pi_1}{\pi_1^2}\frac{t(2-t)}{(1-t)^2}$, with $|f_b'(t)| \leq \frac{1-\pi_1}{\pi_1^2}\frac{1}{(1-t)^2}$ for $0 \leq t \leq 1$, and $f''_b(t) = \frac{1-\pi_1}{\pi_1^2}\frac{2}{(1-t)^3}$, this yields
\begin{align*}
\left|f_b'\left(\E_{Z}\left[h(uz+\sqrt{1-u^2}Z)\right]\right)\right| &\leq \frac{4}{3}C^2\frac{1-\pi_1}{\pi_1^2}e^{2\lambda'z^2}, \\
\left|f_b''\left(\E_{Z}\left[h(uz+\sqrt{1-u^2}Z)\right]\right)\right| &\leq \frac{16}{\sqrt{3}^3}C^3\frac{1-\pi_1}{\pi_1^2}e^{3\lambda'z^2}, \\
\end{align*}
together with the alternative $h \in \mathcal{H}^2_{C,\lambda}$ and $f_a'(t) = 2t$, $f_a''(t) = 2$, this gives
\begin{align*}
\left|f'\left(\E_{Z}\left[h(uz+\sqrt{1-u^2}Z)\right]\right)\right| &\leq \frac{4}{3}C^2\frac{1-\pi_1}{\pi_1^2}e^{2\lambda'z^2} + 2\sqrt{2}Ce^{\lambda z^2}, \\
\left|f''\left(\E_{Z}\left[h(uz+\sqrt{1-u^2}Z)\right]\right)\right| &\leq \frac{16}{\sqrt{3}^3}C^3\frac{1-\pi_1}{\pi_1^2}e^{3\lambda'z^2} + 2.\\
\end{align*}

\paragraph{Proof of 1.(a) and 1.(b).} First, let us show that for any $z$,
\begin{align*}
    \frac{\partial}{\partial u}\left(\E_{Z'}\left[h(uz + \sqrt{1 - u^2}Z')\right]\right) = \E_{Z'}\left[\left(z - \frac{u}{\sqrt{1-u^2}}Z'\right)h'(uz + \sqrt{1-u^2}Z')\right].
\end{align*}
Indeed
\begin{align*}
    \frac{\partial}{\partial u} h(uz + \sqrt{1 - u^2}z') = \left(z - \frac{u}{\sqrt{1-u^2}}z'\right)h'(uz + \sqrt{1-u^2}z')
\end{align*}
which is bounded by $C \cdot (|z| + \frac{|z'|}{\sqrt{1-u_0^2}}) \cdot e^{\lambda(uz + \sqrt{1 - u^2}z')^2}$, which is integrable when replacing $z'$ with $Z'$. Thus, the dominated convergence theorem applies and
\begin{align*}
    \frac{\partial}{\partial u}\left(\E_{Z'}\left[h(uz + \sqrt{1 - u^2}Z')\right]\right) &= \E_{Z'}\left[\frac{\partial}{\partial u}\left(h(uz + \sqrt{1 - u^2}Z')\right)\right]  \\
    &= \E_{Z'}\left[\left(z - \frac{u}{\sqrt{1-u^2}}Z'\right)h'(uz + \sqrt{1-u^2}Z')\right]. \\
\end{align*}
Then, we show that,
\begin{align*}
    &g'(u) \\
    &= \E_Z\left[\E_{Z'}\left[\left(Z - \frac{u}{\sqrt{1-u^2}}Z'\right)h'(uZ + \sqrt{1-u^2}Z')\right] \cdot f'\left(\E_{Z'}\left[h(uZ + \sqrt{1-u^2}Z')\right]\right) \right].
\end{align*}
Indeed, for any $z$,
\begin{align*}
    &\frac{\partial}{\partial u} f\left( \E_{Z'}\left[h(uz + \sqrt{1-u^2}Z')\right] \right) \\
    &=\frac{\partial}{\partial u} \left(\E_{Z'}\left[h(uz + \sqrt{1-u^2}Z')\right]\right) \cdot f'\left(\E_{Z'}\left[h(uz + \sqrt{1-u^2}Z')\right]\right) \\
    &= \E_{Z'}\left[\left(z - \frac{u}{\sqrt{1-u^2}}Z'\right)h'(uz + \sqrt{1-u^2}Z')\right] \cdot f'\left(\E_{Z'}\left[h(uz + \sqrt{1-u^2}Z')\right]\right)
\end{align*}
which is bounded by 
\begin{align*}
    &\E_{Z'}\left[\left(|z| + \frac{1}{\sqrt{1-u_0^2}}|Z'|\right) \cdot Ce^{\lambda(uz + \sqrt{1 - u^2}Z')^2}\right] \cdot f'\left(\E_{Z'}\left[h(uz + \sqrt{1 - u^2}Z')\right]\right)\\
    & \ \ \ \ \ \text{from the assumptions} \\
    &\leq C \left(|z|\sqrt{2}e^{\lambda z^2} + \frac{\sqrt{2}}{\sqrt{1-u_0^2}}(3 + 16\lambda^2z^2)e^{\lambda z^2}\right) \cdot \left(\frac{4}{3}C^2\frac{1-\pi_1}{\pi_1^2}e^{2\lambda'z^2} + 2\sqrt{2}Ce^{\lambda z^2}\right) \\
    & \ \ \ \ \ \text{ from the preliminary bounds,}
\end{align*}
which is integrable when replacing $z$ with $Z$; the dominated convergence theorem gives the result for $g'(u)$. We simplify it further and show it is non-negative. Indeed, $g'(u)$ can be decomposed as
\begin{align*}
    g'(u) = &\E_Z\left[Z \cdot \E_{Z'}\left[h'(uZ + \sqrt{1-u^2}Z')\right] \cdot f'\left(\E_{Z'}\left[h(uZ + \sqrt{1-u^2}Z')\right]\right) \right]\\
    &- \E_Z\left[\frac{u}{\sqrt{1-u^2}} \cdot \E_{Z'}\left[Z'h'(uZ + \sqrt{1-u^2}Z')\right] \cdot f'\left(\E_{Z'}\left[h(uZ + \sqrt{1-u^2}Z')\right]\right) \right].
\end{align*}
Then, note that from Stein's lemma,
\begin{align*}
    \forall z, \E_{Z'}\left[Z'h'(uz + \sqrt{1-u^2}Z')\right] = \sqrt{1-u^2} \cdot \E_{Z'}\left[h''(uz + \sqrt{1-u^2}Z')\right].
\end{align*}
So the second term of the decomposition of $g'(u)$ is equal to
\begin{align*}
    - \E_Z\left[u \cdot \E_{Z'}\left[h''(uZ + \sqrt{1-u^2}Z')\right] \cdot f'\left(\E_{Z'}\left[h(uZ + \sqrt{1-u^2}Z')\right]\right) \right].
\end{align*}
We now turn to the first term of the decomposition. It is equal to
\begin{align*}
    &\E_Z\left[\lim_{R \rightarrow \infty}1_{|Z| \leq R} \cdot Z \cdot \E_{Z'}\left[h'(uZ + \sqrt{1-u^2}Z')\right] \cdot f'\left(\E_{Z'}\left[h(uZ + \sqrt{1-u^2}Z')\right]\right) \right] \\
    &= \lim_{R \rightarrow \infty} \E_Z\left[1_{|Z| \leq R} \cdot Z \cdot \E_{Z'}\left[h'(uZ + \sqrt{1-u^2}Z')\right] \cdot f'\left(\E_{Z'}\left[h(uZ + \sqrt{1-u^2}Z')\right]\right) \right]
\end{align*}
from the dominated convergence theorem as the integrand of the RHS is bounded by $|Z|C\sqrt{2} \cdot e^{\lambda Z^2} \cdot \left(\frac{4}{3}C^2\frac{1-\pi_1}{\pi_1^2}e^{2\lambda'Z^2} + 2\sqrt{2}Ce^{\lambda Z^2}\right)$ which is integrable. Further, we have that
\begin{align*}
    \forall k = 0,1, \ \frac{\partial}{\partial z} \E_{Z'}\left[h^{(k)}(uz + \sqrt{1-u^2}Z')\right] &=  \E_{Z'}\left[\frac{\partial}{\partial z} h^{(k)}(uz + \sqrt{1-u^2}Z')\right]\\
    &= \E_{Z'}\left[uh^{(k+1)}(uz + \sqrt{1-u^2}Z')\right]
\end{align*}
from the dominated convergence theorem as the integrand in the RHS is bounded by $Cue^{\lambda (uR + \sqrt{1-u^2}|Z'|)^2}$ which is integrable. Note that this is where the use of $1_{|Z| \leq R}$ is needed ; otherwise the integrand in the RHS could not be uniformly (wrt $z$) bounded by an integrable function! Thus, we obtain that
\begin{align*}
    &\frac{\partial}{\partial z}\left(\E_{Z'}\left[h'(uz + \sqrt{1-u^2}Z')\right] \cdot f'\left(\E_{Z'}\left[h(uz + \sqrt{1-u^2}Z')\right]\right)\right)\\
    &= u \cdot \bigg(\E_{Z'}\left[h''(uz + \sqrt{1-u^2}Z')\right] \cdot f'\left(\E_{Z'}\left[h(uz + \sqrt{1-u^2}Z')\right]\right)\\
    & \ \ \ \ \ \ \ \ \ \ \ + \E_{Z'}\left[h'(uz + \sqrt{1-u^2}Z')\right]^2f''\left(\E_{Z'}\left[h(uz + \sqrt{1-u^2}Z')\right]\right) \bigg).
\end{align*}
Then, integration by parts gives a slight variation of Stein's lemma, as
\begin{align*}
    &\E_Z\left[1_{|Z| \leq R} \cdot Z \cdot \E_{Z'}\left[h'(uZ + \sqrt{1-u^2}Z')\right] \cdot f'\left(\E_{Z'}\left[h(uZ + \sqrt{1-u^2}Z')\right]\right) \right]\\
    &= \left[-\frac{1}{\sqrt{2\pi}}e^{-z^2/2}\E_{Z'}\left[h'(uz + \sqrt{1-u^2}Z')\right] \cdot f'\left(\E_{Z'}\left[h(uz + \sqrt{1-u^2}Z')\right]\right)\right]_{z = -R}^{z = R} \\
    & \ \ \ + \E_Z\Big[1_{|Z| \leq R} \cdot u \cdot \Big(\E_{Z'}\left[h''(uZ + \sqrt{1-u^2}Z')\right] \cdot f'\left(\E_{Z'}\left[h(uZ + \sqrt{1-u^2}Z')\right]\right)\\
    & \ \ \ \ \ \ \ \ \ \ \ \ \ \ \ \ \ \  \ \ \ \ \ \ \  \ \ \ \ \ \ \ \ \ \ \  \ \ + \E_{Z'}\left[h'(uZ + \sqrt{1-u^2}Z')\right]^2f''\left(\E_{Z'}\left[h(uZ + \sqrt{1-u^2}Z')\right]\right) \Big) \Big]. \\
\end{align*}
The function of $z$ in the $[\dots]_{z = -R}^{z = R}$ brackets is bounded by $\frac{C}{\sqrt{\pi}}e^{\left(\lambda-\frac{1}{2}\right)z^2}\left(\frac{4}{3}C^2\frac{1-\pi_1}{\pi_1^2}e^{2\lambda'z^2} + 2\sqrt{2}Ce^{\lambda z^2}\right)$ which goes to $0$ as $z \rightarrow \pm \infty$, so these $[\dots]_{z = -R}^{z = R}$ brackets vanish when $R \rightarrow \infty$. For the expectation wrt $Z$, note that the integrand is bounded by 
\begin{align*}
&uC\sqrt{2}e^{\lambda Z^2} \left(\frac{4}{3}C^2\frac{1-\pi_1}{\pi_1^2}e^{2\lambda'Z^2} + 2\sqrt{2}Ce^{2\lambda Z^2}\right) + uC^2 \cdot 2e^{2\lambda Z^2} \left(\frac{16}{\sqrt{3}^3}C^3\frac{1-\pi_1}{\pi_1^2}e^{3\lambda'Z^2} + 2 \right)
\end{align*}

which is integrable, so the dominated convergence theorem applies and
\begin{align*}
&\lim_{R \rightarrow \infty}\E_Z\Big[1_{|Z| \leq R} \cdot u \cdot \Big(\E_{Z'}\left[h''(uZ + \sqrt{1-u^2}Z')\right] \cdot f'\left(\E_{Z'}\left[h(uZ + \sqrt{1-u^2}Z')\right]\right)\\
    & \ \ \ \ \ \ \ \ \ \ \ \ \ \ \ \ \ \ \ \ \ \ \ \ \ + \E_{Z'}\left[h'(uZ + \sqrt{1-u^2}Z')\right]^2f''\left(\E_{Z'}\left[h(uZ + \sqrt{1-u^2}Z')\right]\right) \Big) \Big] \\
    &= \E_Z\Big[u \cdot \Big(\E_{Z'}\left[h''(uZ + \sqrt{1-u^2}Z')\right] \cdot f'\left(\E_{Z'}\left[h(uZ + \sqrt{1-u^2}Z')\right]\right)\\
    & \ \ \ \ \ \ \ \ \ \ \ \ \ \ \ \ \ \ \ \ \ \ \ \ \ \ + \E_{Z'}\left[h'(uZ + \sqrt{1-u^2}Z')\right]^2f''\left(\E_{Z'}\left[h(uZ + \sqrt{1-u^2}Z')\right]\right) \Big) \Big], \\
\end{align*}
and the first term of the decomposition of $g'(u)$ is equal to the RHS just above. Thus, summing the first and second terms of the decomposition, the terms involving $h''$ cancel out and we obtain
\begin{align*}
    g'(u) = u \cdot \E_Z\left[\E_{Z'}\left[h'(uZ + \sqrt{1-u^2}Z')\right]^2f''\left(\E_{Z'}\left[h(uZ + \sqrt{1-u^2}Z')\right]\right) \right]
\end{align*}
which is non-negative, concluding the proof for this part. \\[2ex]

\paragraph{Proof of 2.(a)(i).} First note that when $f = f_a$, the above computations can be repeated by replacing $h$ with $h^{(k)}$ for $k = 0, \dots, K$ to show that, noting $\forall k, J_k(u) := \E_Z\left[\E_{Z'}\left[h^{(k)}(uZ + \sqrt{1-u^2}Z')\right]^2 \right]$, every $J_k$ for $k = 0, \dots, K$ is differentiable and, most importantly,
\begin{align*}
    \forall k = 0, \dots, K-1, \ (J_k)'(u) = 2u \cdot J_{k+1}(u).
\end{align*}
Thus, starting from $g(u) = J_0(u)$, we obtain by recursion that $g$ is $K$ times differentiable and for any $ k = 0, \dots, K$, $g^{(k)}(u) = \sum_{i=0}^k p_i^k(u) J_i(u)$ where each $p_i^k(u)$ is a polynomial function of degree at most $i$ with non-negative coefficients, and each $p_k^k(u)$ further has a degree exactly $k$. We note that every $J_i$ and every $p_i^k$ is non-negative, and so is every $p_i^k(.)J_i(.)$, thus every $g^{(k)}$. \\

Further, $J_K(0) > 0$ since $\E_{Z'}\left[h^{(K)}(Z')\right] \neq 0$ by assumption and $J_K$ is continuous as it is differentiable ; thereby $J_K$ is positive on an interval of the form $[0,\epsilon]$ where $\epsilon \in (0,u_0]$. Notably, since $p_K^K$ is of degree exactly $K$ with non-negative coefficients, it is positive on $(0,u_0]$. As a result, $p_K^K(.)J_K(.)$ is positive on $(0,\epsilon]$, thus so is $g^{(K)}$, with $g^{(K)}(0) \geq 0$. This yields $g^{(K-1)}$ being increasing on $[0,\epsilon]$. Remember that it is also non-decreasing on $[\epsilon,u_0]$, since $g^{(K)}$ is generally non-negative. Since $g^{(K-1)}(0) \geq 0$ by non-negativity of $g^{(K-1)}$, we obtain that $\forall u\in(0,\epsilon], \ g^{(K-1)}(u) > g^{(K-1)}(0) \geq 0$ and $\forall u\in[\epsilon,u_0], \ g^{(K-1)}(u) \geq g^{(K-1)}(\epsilon) > g^{(K-1)}(0) \geq 0$, showing that $\forall u\in(0,u_0], \ g^{(K-1)}(u) > 0$ and $g^{(K-1)}(0) \geq 0$. Thus, by recursion we obtain for every $k = K-1, \dots, 1$, $g^{(k)}$ is generally non-negative while being positive on $(0,u_0]$. When $k=1$, this yields that $g$ is an increasing function on $[0,u_0]$, concluding the proof.

\paragraph{Proof of 2.(b).} When $f = f_b$, we still have $f'' \geq 2\frac{1-\pi_1}{\pi_1^2}$ so $g'(u) \geq 2u\frac{1-\pi_1}{\pi_1^2}J_1(u)$, where the RHS is, up to a positive constant, the derivative of $J_0$. Thus, as  $h \in \mathcal{H}'^{K+1}_{C,\lambda,\lambda'} \subset \mathcal{H}^{K+1}_{C,\lambda}$ the proof of 2.(a)(i) can be repeated to show that the derivative of $J_0$ is generally non-negative while being positive on $(0,u_0]$; it immediately follows that $g'$ is also generally non-negative while being positive on $(0,u_0]$. Thus $g$ is an increasing function on $[0,u_0]$. This concludes the proof of 2.(b).

\subsubsection{Proof of 2.(a)(ii)}

 When $h(z) = 1_{\{z \leq z_0\}}$, the normalizing constant is $\mathbb{E}_Z[h(Z)] = \Phi(z_0)$, where $\Phi$ is the CDF of the centered standard Gaussian distribution, and $\densityst{X}(x) = \frac{1_{\{\beta'x \leq z_0\}}}{\Phi(z_0)}$. Then,
\begin{align*}
    \Es\left[\densityst{X}(X) \middle| \gamma'X \right] = \E_{Z \sim \mathcal{N}((\beta'\gamma)\gamma'X, 1 - (\beta'\gamma)^2)}\left[\frac{1_{\{Z \leq z_0\}}}{\Phi(z_0)}\right] = \frac{\Phi\left(\frac{z_0 - (\beta'\gamma)\gamma'X}{\sqrt{1-(\beta'\gamma)^2}}\right)}{\Phi(z_0)}.
\end{align*}
From Formula 20,010.4 of \citet{owen1980atoniat}, we have
\begin{align*}
    \forall a, b, \ \ \E_{Z \sim \mathcal{N}(0,1)}\left[\Phi(a + bZ)^2\right] = \Phi\left(\frac{a}{\sqrt{1 + b^2}}\right) - 2T\left(\frac{a}{\sqrt{1 + b^2}}, \frac{1}{\sqrt{1 + 2b^2}}\right)
\end{align*}
where $T(.,.)$ is Owen's T function. Thus, as $\misoverlapst(\rep_\gamma(X)) = \E_{Z \sim \mathcal{N}(0,1)}\left[\Es\left[\densityst{X}(X) \middle| \gamma'X =Z \right]^2\right]$ from the above, it is equal, up to a $\frac{1}{\Phi(z_0)^2}$ constant, to the RHS of this equation for $a = \frac{z_0}{\sqrt{1-(\beta'\gamma)^2}}$, $b = \frac{-\beta'\gamma}{\sqrt{1-(\beta'\gamma)^2}}$, leading to
\begin{align*}
    \frac{a}{\sqrt{1+b^2}} &= \frac{\frac{z_0}{\sqrt{1-(\beta'\gamma)^2}}}{\sqrt{1 + \frac{(\beta'\gamma)^2}{1-(\beta'\gamma)^2}}} = z_0, \\
    \frac{1}{\sqrt{1 + 2b^2}} &= \frac{1}{\sqrt{1 + 2\frac{(\beta'\gamma)^2}{1-(\beta'\gamma)^2}}} = \sqrt{\frac{1 - (\beta'\gamma)^2}{1 + (\beta'\gamma)^2}} = \sqrt{\frac{2}{1 + (\beta'\gamma)^2} - 1}.
\end{align*}
Thus, in the end,
\begin{align*}
    \misoverlapst(\rep_\gamma(X))= \frac{\Phi(z_0) - 2T\left(z_0, \sqrt{\frac{2}{1 + (\beta'\gamma)^2} - 1}\right)}{\Phi(z_0)^2}.
\end{align*}
As $T$ is increasing in its second argument \citep{owen1980atoniat}, we obtain that $\misoverlapst(\rep_\gamma(X))$ is increasing in $|\beta'\gamma|$.

\subsubsection{Proof of 2.(a)(iii)}
First, noting $\varphi(u) = \frac{1}{\sqrt{2\pi}}e^{-u^2/2}$, for any $\mu$ and any $\sigma > 0$,
\begin{align*}
\mathbb{E}_{Z \sim \mathcal{N}(\mu,\sigma^2)}[\max(0,Z)] 
&= \int_0^\infty z\,\frac{1}{\sigma\sqrt{2\pi}} \exp\left(-\frac{(z-\mu)^2}{2\sigma^2}\right)dz \\[1mm]
&= \int_{-\mu/\sigma}^\infty (\sigma z'+\mu) \frac{1}{\sqrt{2\pi}} e^{-z'^2/2}dz' \\[1mm]
&= \sigma \int_{-\mu/\sigma}^\infty z'\,\frac{1}{\sqrt{2\pi}} e^{-z'^2/2}dz'
+ \mu \int_{-\mu/\sigma}^\infty \frac{1}{\sqrt{2\pi}} e^{-z'^2/2}dz' \\[1mm]
&= \sigma\,\varphi\left(-\frac{\mu}{\sigma}\right)
+ \mu\,\Bigl[1-\Phi\left(-\frac{\mu}{\sigma}\right)\Bigr] \\[1mm]
&= \sigma\,\varphi\left(\frac{\mu}{\sigma}\right)
+ \mu\,\Phi\left(\frac{\mu}{\sigma}\right).
\end{align*}
So the normalizing constant $\E_{Z\sim\mathcal{N}(0,1)}\left[h(Z)\right] = \E_{Z\sim\mathcal{N}(0,1)}\left[\max(0,Z)\right] = \varphi(0) = \frac{1}{\sqrt{2\pi}}$, so $\densityst{X}(x) = \sqrt{2\pi}\max(0,\beta'x)$. Further,
\begin{align*}
    &\Es\left[\densityst{X}(X) \middle| \gamma'X\right] \\
    &= \sqrt{2\pi}\mathbb{E}_{Z \sim \mathcal{N}((\beta'\gamma)\gamma'X, 1 - (\beta'\gamma)^2)}[\max(0,Z)] \\
    &= \sqrt{2\pi}\left(\sqrt{1 - (\beta'\gamma)^2}\varphi\left(\frac{(\beta'\gamma) \gamma'X}{\sqrt{1 - (\beta'\gamma)^2}}\right) + (\beta'\gamma) \gamma'X\Phi\left(\frac{(\beta'\gamma) \gamma'X}{\sqrt{1 - (\beta'\gamma)^2}}\right)\right).
\end{align*}
Thus, noting $u = \beta'\gamma$, and $a = \frac{u}{\sqrt{1-u^2}}$,
\begin{align*}
    \misoverlapst(\rep_\gamma(X))
    &= \E_{Z \sim \mathcal{N}(0,1)}\left[\Es\left[\densityst{X}(X) \middle| \gamma'X = Z\right]^2\right] \\
    &= 2\pi\E_{Z \sim \mathcal{N}(0,1)}\left[\left(\sqrt{1 - u^2}\varphi\left(\frac{uZ}{\sqrt{1 - u^2}}\right) + uZ\Phi\left(\frac{uZ}{\sqrt{1 - u^2}}\right)\right)^2\right] \\
    &= 2\pi(1-u^2)\E_{Z \sim \mathcal{N}(0,1)}\left[\varphi(aZ)^2\right] \\
    & \ \ \ \ \ \ \ \ \ \ + 4\pi u\sqrt{1-u^2}\E_{Z \sim \mathcal{N}(0,1)}\left[Z\varphi(aZ)\Phi(aZ)\right]\\
    & \ \ \ \ \ \ \ \ \ \ + 2\pi u^2\E_{Z \sim \mathcal{N}(0,1)}\left[Z^2\Phi(aZ)^2\right]
\end{align*}
where
\begin{align*}
    \E_{Z \sim \mathcal{N}(0,1)}\left[\varphi(aZ)^2\right] &= \int \frac{1}{2\pi}e^{-a^2z^2}\frac{1}{\sqrt{2\pi}}e^{-z^2/2}dz \\
    &= \frac{1}{\sqrt{2\pi}^3} \int e^{-(2a^2+1)z^2/2} dz \\
    &= \frac{1}{2\pi\sqrt{2a^2+1}} \\
    &= \frac{1}{2\pi\sqrt{\frac{2u^2}{1-u^2}+1}} \\
    &= \frac{1}{2\pi}\sqrt{\frac{1-u^2}{1+u^2}}
\end{align*}
and, from Stein's lemma,
\begin{align*}
     &\E_{Z \sim \mathcal{N}(0,1)}\left[Z\varphi(aZ)\Phi(aZ)\right]\\
     &= \E_{Z \sim \mathcal{N}(0,1)}\left[a\varphi'(aZ)\Phi(aZ) + a\varphi(aZ)\Phi'(aZ)\right] \\
     &= -a^2\E_{Z \sim \mathcal{N}(0,1)}\left[Z\varphi(aZ)\Phi(aZ)\right] + a\E_{Z \sim \mathcal{N}(0,1)}\left[\varphi(aZ)^2\right]
\end{align*}
so
\begin{align*}
    \E_{Z \sim \mathcal{N}(0,1)}\left[Z\varphi(aZ)\Phi(aZ)\right] &= \frac{a}{1+a^2}\E_{Z \sim \mathcal{N}(0,1)}\left[\varphi(aZ)^2\right] \\
    &= \frac{\frac{u}{\sqrt{1-u^2}}}{1+\frac{u^2}{1-u^2}}\frac{1}{2\pi}\sqrt{\frac{1-u^2}{1+u^2}} \\
    &= \frac{u}{2\pi}\frac{1-u^2}{\sqrt{1+u^2}}.
\end{align*}
Thus, again from Stein's lemma, and from Formula 2,010.3 of \citet{owen1980atoniat},
\begin{align*}
    &\E_{Z \sim \mathcal{N}(0,1)}\left[Z^2\Phi(aZ)^2\right]\\
    &= \E_{Z \sim \mathcal{N}(0,1)}\left[\Phi(aZ)^2 + 2aZ\Phi'(aZ)\Phi(aZ)\right] \\
    &= \E_{Z \sim \mathcal{N}(0,1)}\left[\Phi(aZ)^2\right] + 2a\E_{Z \sim \mathcal{N}(0,1)}\left[Z\varphi(aZ)\Phi(aZ)\right]  \\
    &= \frac{1}{4} + \frac{1}{2\pi}\arcsin\left(\frac{a^2}{1+a^2}\right) + 2a\frac{u}{2\pi}\frac{1-u^2}{\sqrt{1+u^2}} \\
    &= \frac{1}{4} + \frac{1}{2\pi}\arcsin\left(\frac{\frac{u^2}{1-u^2}}{1+\frac{u^2}{1-u^2}}\right) + 2\frac{u}{\sqrt{1-u^2}}\frac{u}{2\pi}\frac{1-u^2}{\sqrt{1+u^2}} \\
    &= \frac{1}{4} + \frac{1}{2\pi}\arcsin\left(u^2\right) + \frac{u^2}{\pi}\sqrt{\frac{1-u^2}{1+u^2}}. \\
\end{align*}
In the end,
\begin{align*}
    &\misoverlapst(\rep_\gamma(X))\\
    &= 2\pi(1-u^2)\frac{1}{2\pi}\sqrt{\frac{1-u^2}{1+u^2}} + 4\pi u\sqrt{1-u^2}\frac{u}{2\pi}\frac{1-u^2}{\sqrt{1+u^2}}\\
    & \ \ \ \ \ \ \ \ \ \ + 2\pi u^2\left(\frac{1}{4} + \frac{1}{2\pi}\arcsin\left(u^2\right) + \frac{u^2}{\pi}\sqrt{\frac{1-u^2}{1+u^2}}\right) \\
    &= \frac{\sqrt{1-u^2}^3}{\sqrt{1+u^2}} + \frac{2u^2\sqrt{1-u^2}^3}{\sqrt{1+u^2}} + \frac{\pi}{2}u^2 + u^2\arcsin(u^2) + \frac{2u^4\sqrt{1-u^2}}{\sqrt{1+u^2}} \\
    &= \sqrt{\frac{1-u^2}{1+u^2}}\cdot\left(1-u^2 + 2u^2(1-u^2) + 2u^4\right) + \frac{\pi}{2}u^2 + u^2\arcsin(u^2) \\
    &=  \sqrt{\frac{1-u^2}{1+u^2}}\cdot\left(1+u^2\right) + \frac{\pi}{2}u^2 + u^2\arcsin(u^2) \\
    &=  \sqrt{1-w^2} + \frac{\pi}{2}w + w\arcsin(w) \\
\end{align*}
where $w := u^2 = (\beta'\gamma)^2$, and $\misoverlapst(\rep_\gamma(X))$ is an increasing function of $w$ as
\begin{align*}
    \frac{\partial}{\partial w}\misoverlapst(\rep_\gamma(X)) = \frac{-2w}{2\sqrt{1-w^2}} + \frac{\pi}{2} +\arcsin(w) + \frac{w}{\sqrt{1-w^2}} = \frac{\pi}{2} +\arcsin(w)
\end{align*}
which is positive as $w \geq 0$. This concludes the proof.

\section{FURTHER DISCUSSION ON PREVIOUS WORK AND THEORETICAL RESULTS}

\subsection{Comparison with Approaches Inspired by Domain Adaptation}
\label{sec:comparison_with_domain_adaptation_approaches}

When comparing our approach to the related works inspired by domain adaptation, as in  \citet{shalit2017eitegbaa}, \citet{johansson2022gbarlfeopoace} or \citet{zhang2020lorfteoite}, one subtlety is that the appropriate metric for measuring overlap may vary by (1) the target of estimation, and (2) the parametric assumptions one is willing to impose on the outcome process. In our approach, we target the scalar ATT, while the aforementioned related works target the Conditional Average Treatment Effect (CATE), which is a whole function. As a result, while our work leverages the semiparametric efficiency bound of the ATT as a scalar objective, the latter cannot readily be applied to the CATE. Instead, the related works construct bounds under different sets of assumptions. Each of the related works decomposes the MSE of the CATE into a factual, observable part and a counterfactual, unobservable part. The unobservable part is then further upper-bounded by a quantity that measures overlap, but the particular overlap metric depends on parametric restrictions on the unknown counterfactual outcome model. \citet{shalit2017eitegbaa} and \citet{johansson2022gbarlfeopoace} assume that the pointwise loss of the counterfactual regressor lives in a restricted function class, so use an IPM to bound the counterfactual loss, while \citet{zhang2020lorfteoite} assume a posterior measure over the counterfactual regressor, which makes the posterior variance relevant for PAC-Bayes bounds.

Further, note that some works such as \citet{assaad2021crlwbw} and \citet{ johansson2022gbarlfeopoace} incorporate inverse probability weights in both the observable part of the MSE of the CATE and the quantity measuring overlap. This can improve the estimation of the CATE as \citet{assaad2021crlwbw} shows that introducing such weights reduces the contradiction between the minimizations of these two components, while \citet{johansson2022gbarlfeopoace} notes that the resulting reweighted overlap measure should be lower than the original one. However, this approach is not directly applicable to our setup where we aim to find representations with better overlap \textit{in the original distribution}. Indeed, the reweighted overlap measure measures the overlap between treated and control representation \textit{reweighted} distributions and minimizing it does not encourage learning representations with better overlap in the original distribution, especially as the weights estimate true inverse propensity weights which render treated and control distributions of both original covariates and \emph{any} representation identical by design.

\subsection{Lemma \ref{lem:cb}}
\label{app:discussion_cb}
  Lemma \ref{lem:cb} gives a ``doubly robust'' moment that identifies deconfounding scores. This is analogous to the ``mixed bias property'' in automatic debiased machine learning (AutoDML) for sensitivity analysis \citep{chernozhukov2022rafadmlwnnarf}, where the bias of the estimator is the correlation of the residuals from an outcome model regression and a Riesz representer regression. In our setting, each of these regressions would be the perfect regression of its corresponding ground-truth model on the representation $\rep(X)$. \citet{chernozhukov2022lssovbicml} established a mixed bias property for AutoDML in sensitivity analysis; our Lemma \ref{lem:cb} is a special case for the ATT and adapted to our broader setup.
  
\subsection{Lemma \ref{lem:misoverlap}}
\label{app:discussion_misoverlap}

Lemma \ref{lem:misoverlap} shows that overlap divergence is a proxy for the semiparametric efficiency bound of the estimand obtained by adjusting on the representation; this motivates our specific choice of the $\chi^2$-divergence.

 In general, the assumptions on outcomes in Items 1 and 2 are relatively mild and commonly invoked in the literature: the outcome is binary in many applications, e.g. medicine \citep{colnet2023rrorrdwcmietg}, and $\text{Var}_{\Ps}(Y|X=x)$ is constant in several popular treatment effect estimation datasets such as IHDP \citep{hill2011bnmfci} or News \citep{johansson2016lrfci}.

\subsection{Theorem \ref{th:hyperbola}}
\label{app:discussion_hyperbola}

When $\alpha'\beta < 0$, the deconfounding score with coordinate $w_1$ on the segment with prognostic score endpoint $\alpha$  in $\mathcal{D}_{\alpha, \beta}$ is equal to the deconfounding score with coordinate $w_2$ on the segment with prognostic score endpoint $\alpha$  in $\mathcal{D}_{\alpha, -\beta}$. The same statement can be made when replacing $\alpha$ with $-\alpha$ as the prognostic score endpoint. More generally, regardless of the sign of $\alpha'\beta$, Assumptions \ref{ass:lineargaussian1} and \ref{ass:lineargaussian2} can be expressed equivalently when replacing $\beta$ with $-\beta$ and $h$ with $h(-.)$, and any $\gamma$ with coordinates $(w_1,w_2)$ and orthogonal component $n$ in $\mathcal{D}_{\alpha, \beta}$ can be expressed with coordinates $(w_2,w_1)$ and the same orthogonal component $n$ in $\mathcal{D}_{\alpha, -\beta}$.

\subsection{Theorem \ref{th:misoverlap_prognostic_score}}
\label{app:discussion_misoverlap_prognostic_score}

The assumptions besides those imposing Gaussian covariates and generalized linear models are not too stringent: the upper bounds in the classes $\mathcal{H}^K_{C,\lambda}$ and  $\mathcal{H}'^K_{C,\lambda,\lambda'}$ simply ensure two-factor products of these functions remain integrable for the standard centered Gaussian density measure, and in particular that the overlap divergence of $X$ is finite in Item 1.(a) of Theorem \ref{th:misoverlap_prognostic_score}, in line with Assumption \ref{ass:strong_overlap}. The lower bound on $1 - h(z)$ in $\mathcal{H}'^K_{C,\lambda,\lambda'}$ can be interpreted as ensuring that $e(X)$ does not converge to 1 too fast when $X$ takes large values, a condition that is reasonable if one wants sufficient overlap wrt $X$. Notably, it is verified for the classical case $h(z) = \text{logit}^{-1}(z)$ for any $\lambda' > 0$ if one takes sufficiently high $C$. However it is perhaps the most stringent condition in Theorem \ref{th:misoverlap_prognostic_score} as one could allow up to $\lambda' < \frac{1}{2}$ to allow a finite overlap divergence in $X$, while the $\lambda' < \frac{1-4\lambda}{6}$ assumption implies $\lambda' < \frac{1}{6}$. The additional condition of some derivative of $h$ admitting a non-zero expectation for the standard centered Gaussian density measure in Items 2.(a)(i) and 2.(b) of Theorem \ref{th:misoverlap_prognostic_score} is also not stringent, e.g. it is again verified for $h(z) = \text{logit}^{-1}(z)$. Items 2.(a)(ii) and 2.(a)(iii) show that the conclusions of Theorem \ref{th:misoverlap_prognostic_score} also hold for at least some non-continuously-differentiable functions; notably, $h = \text{ReLU}$ is a popular link function in the machine learning community.

Further, note that for any $K, C, \lambda, \lambda'$, if $h \in \mathcal{H}^K_{C,\lambda}$ then $h(-.) \in \mathcal{H}^K_{C,\lambda}$, if $h \in \mathcal{H}'^K_{C,\lambda,\lambda'}$ then $h(-.) \in \mathcal{H}'^K_{C,\lambda,\lambda'}$, and if $\E_{Z \sim \mathcal{N}(0,1)}\left[h^{(K)}(Z)\right] \neq 0$ then $\E_{Z \sim \mathcal{N}(0,1)}\left[h(-.)^{(K)}(Z)\right] \neq 0$, justifying the operation replacing $\beta$ with $-\beta$ and $h$ with $h(-.)$ when $\alpha'\beta < 0$ as the assumptions of Items 1.(a), 1.(b), 2.(a)(i) and 2.(b) of Theorem \ref{th:misoverlap_prognostic_score} will still hold even after this transformation.

\subsection{Potential Extensions}
\label{app:extensions}

While we focused on the ATT, our results can be immediately extended to more general covariate shift, thus to other causal inference problems such as estimation of the classical average treatment effect (ATE) or transportability \citep{clivio2024trlfwpidbci}.

Note that our work straightforwardly extends to non-standard Gaussian variables: if $X \sim \mathcal{N}(0, \Sigma)$ then the entire Section  \ref{sec:linear_gaussian} is valid by replacing $X, \alpha, \beta$ with $\Sigma^{-1/2}X$, $\Sigma^{1/2}\alpha$, $\Sigma^{1/2}\beta$, respectively. Computing closed-form deconfounding scores with any further relaxations in assumptions will be difficult as our results generally require computing the distribution of $f(X)$ given $\rep(X)$, which to the best of our knowledge is classically known only for linear $f, \rep$ and Gaussian $X$. For deconfounding scores, only the equivalence between independence and non-correlation for Gaussian covariates allows GLM forms of $f$. However, a potential direction would be to assume independent Poisson covariates and $\rep_B(X) = \sum_{i \in B}X_i$ for some $B \subset \{1, \dots, d\}$, as the distribution of $(X_j)_{j \in B}$ conditional on $\rep_B(X)$ is known to be multinomial with parameters depending on individual Poisson parameters \citep{townes2020reviewprobabilitydistributionsmodeling}. Additionally, if results could be generalized to multivariate linear representations, then the GLM assumption would encompass widely-used neural networks.

We chose the one-dimensional class of representation in this paper because it could be characterized analytically, and so that we could develop intuition for the unbiasedness constraint implied by Lemma \ref{lem:cb}. While representations from causal deep learning \citep{shalit2017eitegbaa, johansson2022gbarlfeopoace, zhang2020lorfteoite} are able to estimate richer sets of prognostic/balancing scores, deconfounding scores are designed to have zero confounding bias and their overlap divergence directly controls the semiparametric efficiency bound. How to bridge these two types of representations remains an open question. In one direction, Assumptions \ref{ass:lineargaussian1} and \ref{ass:lineargaussian2} could be relaxed to incorporate neural-network outcome and propensity models. In another direction, many of the insights in our paper can be applied to designing regularizers that (1) enforce the zero confounding bias constraint as in Lemma \ref{lem:cb}, and (2) incorporate the overlap divergence from our paper. Note that this proposal actually addresses one of the key questions in \citet{johansson2022gbarlfeopoace} about how regularization should be chosen for causal inference. Generally, we expect the resulting representations to outperform those from the current causal deep learning literature, as they will be both flexible and suited to preserving unconfoundedness and improving overlap. Again, we view this as important motivation for our own work, as it lays the groundwork for such representations.

\section{DETAILS ON EXPERIMENTS}
\label{app:xps}

\paragraph{ATT Estimators.} Here we present the analogs of the IPW \citep{horvitz1952agoswrfafu} and AIPW \citep{robins1994eorcwsranao} estimators of the ATT \citep{moodie2018adrweotateott}; using estimators $\hat{e}(X)$ of $e(X)$ and $\hat m_0(X)$ of $m_0(X)$:
\begin{align}
\label{eq:IPW_ATT}
\hat \tau_{IPW}^{ATT} :=\frac{\sum_{i=1}^N T_iY_i}{\sum_{i=1}^N T_i} - \frac{\sum_{i=1}^N (1-T_i)\frac{\hat{e}(X_i)}{1-\hat{e}(X_i)}Y_i}{\sum_{i=1}^N T_i} 
\end{align}
and 
\begin{align}
\label{eq:AIPW_ATT}
&\hat \tau_{AIPW}^{ATT}
:=  %
\frac{\sum_{i=1}^N T_i(Y_i-\hat{m}_0(X_i))}{\sum_{i=1}^N T_i} - \frac{\sum_{i=1}^N (1-T_i)\frac{\hat{e}(X_i)}{1-\hat{e}(X_i)}(Y_i-\hat{m}_0(X_i))}{\sum_{i=1}^N T_i} .
\end{align}

Further, applying Hajek normalization \citep{hajek1971coapbdb} gives the following estimators, which we use in our simulations:
\begin{align}
\label{eq:IPW_ATT_Hajek}
\hat\tau_{IPW}^{ATT,\text{Hajek}} =\frac{\sum_{i=1}^N T_iY_i}{\sum_{i=1}^N T_i} - \frac{\sum_{i=1}^N (1-T_i)\frac{\hat{e}(X_i)}{1-\hat{e}(X_i)}Y_i}{\sum_{i=1}^N (1-T_i)\frac{\hat{e}(X_i)}{1-\hat{e}(X_i)}} 
\end{align}
and 
\begin{align}
\label{eq:AIPW_ATT_Hajek}
\hat \tau_{AIPW}^{ATT,\text{Hajek}} :=
\frac{\sum_{i=1}^N T_i(Y_i-\hat{m}_0(X_i))}{\sum_{i=1}^N T_i} - \frac{\sum_{i=1}^N (1-T_i)\frac{\hat{e}(X_i)}{1-\hat{e}(X_i)}(Y_i-\hat{m}_0(X_i))}{\sum_{i=1}^N (1-T_i)\frac{\hat{e}(X_i)}{1-\hat{e}(X_i)}}.
\end{align}

\paragraph{Trimming.} To avoid division by zero $1-\hat{e}(X_i)$ in the above estimators, we apply the following transformation to the original propensity score estimator $\hat{e}_O$:
\begin{align*}
    \hat{e}(X_i) = \begin{cases}
    1-\epsilon & \text{if $1-\hat{e}_O(X_i)<\epsilon$},\\
    \hat{e}_O(X_i) & \text{otherwise},
  \end{cases}
\end{align*}
where $\epsilon$ is chosen at R's machine epsilon value \texttt{.Machine\$double.eps}, equal to $2.220446 \times 10^{-16}$ \citep{rcoreteam2024r}.

\paragraph{Generation of $\alpha$ and $\beta$ in Simulated Datasets.} We fix a support $\mathcal{S} \subset \{1, \dots, p\}$ for both $\alpha$ and $\beta$. To generate $\alpha$, we generate $\bar{\alpha}$ as $\forall i \in \mathcal{S}, \bar{\alpha}_i \sim \mathcal{N}(0,1)$, and $\forall i \notin \mathcal{S}, \bar{\alpha}_i = 0$, then retrieve $\alpha = \frac{\bar{\alpha}}{||\bar{\alpha}||}$. We then generate $\beta$ with support $\mathcal{S}$ and chosen such that $\alpha'\beta = K$ for fixed $K \in (-1,1)$ as follows. Again, we generate $u$ such that $\forall i \in \mathcal{S}, u_i \sim \mathcal{N}(0,1)$, and $\forall i \notin \mathcal{S}, u_i = 0$. We construct $v$ as $v = u - (\alpha'u)\alpha$, which is the canonical Gram-Schmidt vector and is orthogonal to $\alpha$ by construction, and deduce $v_n = \frac{v}{||v||}$. Finally, we take $\beta = K\alpha + \sqrt{1-K^2}v_n$. Throughout simulations, we chose $\mathcal{S} = \{1, \dots, 20\}$ and $K = 0.75$. We sampled $\alpha$ and $\beta$ with the same seed across all simulations; notably the values of $\alpha$ in $\mathcal{S}$ were $0.283645181$, $ -0.073268306$, $ 0.298657804$, $ 0.285773167$, $ 0.093123755$, $ -0.345855281$, $ -0.208545604$, $ -0.066190863$, $ -0.001295241$, $ 0.540057827$, $ 0.171494414$, $ -0.179448398$, $ -0.257750723$, $ -0.065009780$, $ -0.067200315$, $ -0.092420657$, $ 0.056646520$, $ -0.200315350$, $ 0.097849518$, $ -0.277937067$ and those of $\beta$ in $\mathcal{S}$ were $0.15135872$, $ 0.02785895$, $ 0.23680749$, $ 0.36840029$, $ 0.05315568$, $ -0.13631776$, $ 0.08172226$, $ -0.19110589$, $ -0.27014375$, $ 0.38747427$, $ 0.07053042$, $ -0.23927785$, $ -0.27107102$, $ -0.18158571$, $ 0.10532350$, $ 0.17679514$, $ 0.24756079$, $ -0.23010864$, $ 0.32796449$, $ -0.25291885$.

\paragraph{Zero Estimated $\hat{\alpha}$ and $\hat{\beta}$.} Deconfounding scores are ill-defined when either $\alpha$ or $\beta$ is zero. However, in this scenario, both adjusting for $X$ and adjusting for an empty variable $Z = \emptyset$ yield the same correct ATT estimand $\tau$, so we conjecture that there is no motivation for even using a deconfounding score. As a result, when we find either $\hat{\alpha}$ or $\hat{\beta}$ to be (near-)zero, we impute deconfounding score estimates to be identical to original covariate estimates. We apply this rule when $||\hat{\alpha}_O||_\infty < 10^{-10}$ or $||\hat{\beta}_O||_\infty < 10^{-10}$, where $\hat{\alpha}_O, \hat{\beta}_O$ are the original unnormalized coefficient vectors obtained from the treatment assignment and outcome model regressions, respectively. This remains a minor phenomenon, however. In simulated datasets, this only took place for 4 out of 16 hyperparameter assignments, and within those for at most 3 out of 100 draws of the data. In ACIC2016 datasets, this only took place for 75 out of 308 hyperparameter assignments, and within them for at most 6 out of 100 draws of the data. IHDP and HC-MNIST were unaffected by this phenomenon. We refer to the \texttt{\_warnings.txt} files produced by the code in the \texttt{results} folder.

\paragraph{Zero-Variance Deconfounding Scores.} It remains possible that estimated deconfounding scores have either zero empirical variance, rendering both propensity score and outcome model estimation impossible, or zero control empirical variance, rendering outcome model estimation impossible. In this case, again, we impute estimates for this specific deconfounding score to be identical to original covariate estimates. This phenomenon only took place for one deconfounding score in the entire experiments, the equiangular score in the 67th iteration of ACIC 2016 with setting 47 and LASSO estimators.

\paragraph{Ground-Truth ATT in Semi-Synthetic Datasets.} In semi-synthetic datasets, covariates are based on real-world studies but outcome (and most often treatment) models are given and synthetic. Thus, such datasets provide $m_t(X_i)$ for all $i = 1, \dots, N$ and $t = 0, 1$ but not the ATT due to the unknown distribution $\Pall_X$; thus, we set the ground-truth ATT to $\frac{\sum_{i=1}^{N}T_i\Delta m(X_i)}{\sum_{i=1}^{N}T_i}$.

\paragraph{Source and Settings in Datasets.} IHDP \citep{hill2011bnmfci} was implemented according to the \texttt{npci} package available at \url{https://github.com/vdorie/npci}, specifically from the \texttt{generateDataForIterInCurrentEnvironment} function available in the package's IHDP example. For that function, we chose \texttt{w = 0.5} and all covariates. We varied the IHDP setting between A, B, C and overlap between lower overlap and higher overlap; this gives 6 settings. ACIC 2016 was implemented using its official implementation of \citet{dorie2017avdiymfcillfadac}, available at \url{https://github.com/vdorie/aciccomp}; its 77 settings are described in \citet{dorie2017avdiymfcillfadac}. HC-MNIST \citep{jesson2021qiiilceeuhc} was converted from its original Python implementation available at \url{https://github.com/anndvision/quince/blob/main/quince/library/datasets/hcmnist.py} to R, with default parameters, and $\Gamma^\star = 1$ to remove the influence of unobserved confounders.

\paragraph{Infrastructure.} Experiments were conducted on a single CPU, of model name ``Intel(R) Core(TM) i7-8750H CPU @ 2.20GHz CPU'' with 2 threads per core, 6 cores per socket, and 1 socket. %

\end{document}